\documentclass{article}

\PassOptionsToPackage{numbers, sort&compress, square}{natbib}

\usepackage[eandd,final]{neurips_2026}

\setcitestyle{numbers}
\usepackage[T1]{fontenc}    
\usepackage{hyperref}       
\usepackage{url}      
\usepackage{booktabs}       
\usepackage{amsfonts}       
\usepackage{nicefrac}       
\usepackage{microtype}      
\usepackage{xcolor}         
\usepackage{amsmath}
\usepackage{amssymb}
\usepackage{float}
\usepackage{tcolorbox}
\usepackage{graphicx}
\usepackage{subcaption}
\usepackage{mathtools}
\usepackage{enumitem}
\usepackage{subcaption}
\usepackage{booktabs}
\usepackage{multirow}
\usepackage{makecell}
\usepackage{caption}
\usepackage{xspace}
\usepackage{makecell}
\usepackage{colortbl}
\usepackage{subcaption}

\usepackage{wrapfig}

\definecolor{bestgreen}{RGB}{220,245,220}
\definecolor{secondred}{RGB}{252,228,214}
\usepackage{xcolor}

\definecolor{violetbench}{RGB}{180, 120, 255}   
\definecolor{greenbench}{RGB}{120, 200, 120}    
\definecolor{yellowbench}{RGB}{255, 200, 80}    

\newcommand{\algo}{\textsc{PURGE}}

\newcommand{\best}[1]{\cellcolor{bestgreen}\textbf{#1}}
\newcommand{\second}[1]{\cellcolor{secondred}#1}

\usepackage{pifont}
\newcommand{\one}{\ding{182}~\!}
\newcommand{\two}{\ding{183}~\!}
\newcommand{\three}{\ding{184}~\!}

\newcommand{\RA}{\ding{228}\hspace{0.2em}}
\usepackage{float}
\title{Partition-Aware Unlearning for Removing Spurious Correlations in Large Vision-Language Models}

\author{
\textbf{Aditi Sarker$^{\dagger}$} \qquad
\textbf{Nazreen Shah$^{\S}$} \qquad
\textbf{Rafi Ibn Sultan$^{\dagger}$} \qquad \textbf{Rhongho Jang$^{\dagger \ddagger}$}\\
\textbf{Dongxiao Zhu$^{\dagger \ddagger}$} \qquad
\textbf{Prashant Khanduri$^{\dagger \ddagger}$\thanks{Corresponding author, correspondence to
\texttt{khanduri.prashant@wayne.edu}}}
\vspace{2 mm} \\
{\small $^{\dagger}$Department of Computer Science, Wayne State University, MI, USA}\\
{\small $^{\S}$Department of Electronics \& Communications Engineering (ECE), IIIT Delhi, India }\\
{\small $^{\ddagger}$Institute for AI and Data Science (AIDaS), Wayne State University, MI, USA  
}
}

\begin{document}
\maketitle


\begin{abstract}
Large Vision-Language Models (LVLMs) achieve strong performance across many multimodal tasks; however, they often exploit spurious object-background correlations, resulting in predictions driven by contextual shortcuts rather than object-relevant visual evidence. Despite growing interest in hallucination and robustness evaluation, existing benchmarks provide limited control over whether model predictions are grounded in the target object or induced by correlated background cues. In this work, we introduce {\bf \algo}~(\underline{P}artition-aware \underline{U}nlearning for \underline{R}emoving spurious-correlation \underline{G}enerated \underline{E}rrors), a framework for constructing, benchmarking, and mitigating spurious-correlation-induced failures in LVLMs. The framework consists of: -- \one {\em Structured dataset construction} wherein we develop three complementary structured data construction strategies that partition examples by object-relevant evidence and spurious background cues, enabling controlled diagnosis of shortcut reliance; and -- \two {\em Partition-aware unlearning}, which uses these partitions to selectively remove spurious object-background associations while preserving object-based reasoning. We evaluate the \algo~framework across multiple LVLMs, including LLaVA-1.6-7B, Qwen3-VL-8B-Instruct, and Qwen3.5-9B, together with CLIP as a vision-language encoder, on a diverse suite of benchmarks, including CHAIR, POPE, Causal-HalBench, MM-SpuBench, AMBER, MMHal, and Waterbirds. Our results show that \algo~consistently reduces hallucinations and spurious-correlation-driven errors while maintaining or improving
overall performance in most evaluated settings, providing both a reusable evaluation protocol and an effective mitigation framework for more reliable LVLMs. PURGE dataset and code are available: \href{https://huggingface.co/datasets/Krishnaaditi1/PURGE}{Dataset (Hugging Face)} \;|\; \href{https://github.com/krishnaaditi/PURGE}{Code (GitHub)}
\end{abstract}

\section{Introduction and Motivation}
\label{intro}
\vspace{-1mm}

Large Vision-Language Models (LVLMs) have achieved strong performance on
multimodal tasks such as visual question answering, image captioning, and
visual reasoning~\cite{li2023blip, liu2023visual, zhu2023minigpt,
dai2023instructblip, wang2024cogvlm, DBLP:journals/corr/abs-2511-21631}.
Despite these advances, LVLMs remain prone to hallucination, where models
generate or predict objects that are not present in the image~\cite{rohrbach2018object,
wang2023amber, biten2022let}. Such failures limit their reliability in
safety-critical and high-stakes applications, including assistive technologies,
autonomous driving, and medical image understanding, where predictions must be
grounded in visual evidence rather than contextual shortcuts. One important
driver of these failures is the presence of \emph{spurious correlations} in
training data~\cite{geirhos2020shortcut, torralba2011unbiased,
sagawa2019distributionally}, where objects frequently co-occur with particular
backgrounds, scenes, or contextual cues.

\begin{figure}[t]
    \centering
    \includegraphics[width=0.65\linewidth]{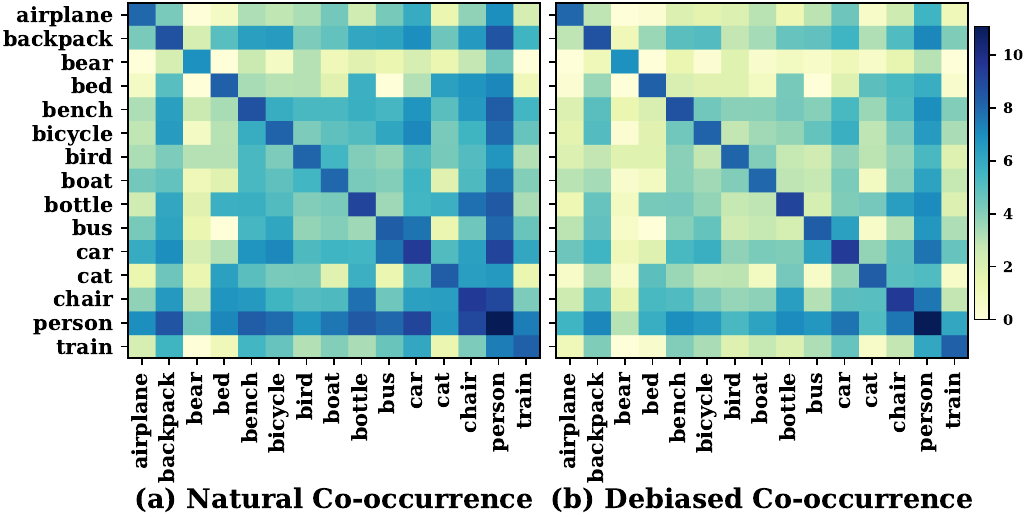}
      \vspace{-2mm}
 
    \caption{\small
    \textbf{Spurious correlations in LVLMs.}
    (a) Natural co-occurrence statistics: Strong object-background dependencies.
    (b) Debiased co-occurrence: Suppresses these dependencies, reducing reliance on spurious correlations.}
    \vspace{-3mm}
    \label{fig:cooccur_heatmap}
\end{figure}

\begin{wrapfigure}{r}{0.53\textwidth}
    \centering
\vspace{-1mm}
    \includegraphics[width=0.80\linewidth]{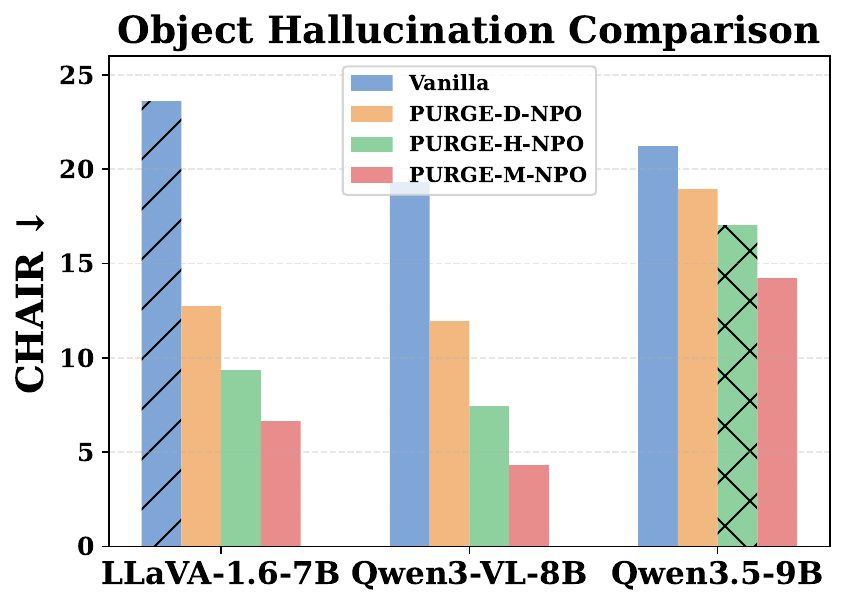}
     \vspace{-2mm}
    \caption{
    Object hallucination comparison measured using CHAIR$_i$  ($\downarrow$); lower values indicate fewer hallucinations.}
    \label{fig:chair_intro}
\end{wrapfigure}

Existing approaches attempt to mitigate this problem through representation learning and multimodal alignment~\cite{yang2023mitigating, lu2025mitigating, jung2024unified}, region-level supervision~\cite{varma2024ravl}, causal or invariant learning~\cite{arjovsky2019invariant, song2024learning, hu2025causal}, and inference-time decoding or prompting strategies~\cite{wang2024mitigating, leng2024mitigating, huang2024opera}. While these methods improve model behavior, they provide limited mechanisms for explicitly identifying and removing spurious object-background associations learned by LVLMs. Training-based methods may still inherit biases from correlated data distributions, whereas inference-time methods adjust model outputs without directly modifying the underlying learned associations. As a result, LVLMs may continue to rely on background shortcuts even when the target object evidence is absent, ambiguous, or contradicted. A central challenge is that existing evaluations often measure hallucination or robustness at the aggregate level, without systematically disentangling whether a prediction is supported by object-relevant visual evidence or induced by correlated background cues.  For example, in real-world datasets such as MSCOCO~\cite{lin2014microsoft}, objects frequently appear in characteristic contexts, creating strong object-background co-occurrence patterns. As shown in Fig.~\ref{fig:cooccur_heatmap}a, such correlations are highly visible in the original data distribution and can encourage models to associate objects with backgrounds rather than with object-specific evidence. In contrast, Fig.~\ref{fig:cooccur_heatmap}b illustrates a debiased distribution with reduced object--background co-occurrences, which translates into improved performance as shown later in Fig.~\ref{fig:chair_intro}. This motivates the key idea of our work:

\begin{tcolorbox}[title=Key idea,
    before skip=3 pt,
    after skip=3 pt]
  \vspace{-2 mm}
  Spurious object-background correlations can be diagnosed and mitigated through {\bf structured data partitioning}. By separating object-grounded examples from shortcut-prone ones, we obtain both a controlled measure of background reliance and targeted retain/forget partitions for {\bf LVLM unlearning}, enabling the model to suppress spurious associations while preserving object-based reasoning.
  \vspace{-2 mm}
\end{tcolorbox}

\begin{figure}[H]
\centering
\includegraphics[width=\linewidth]{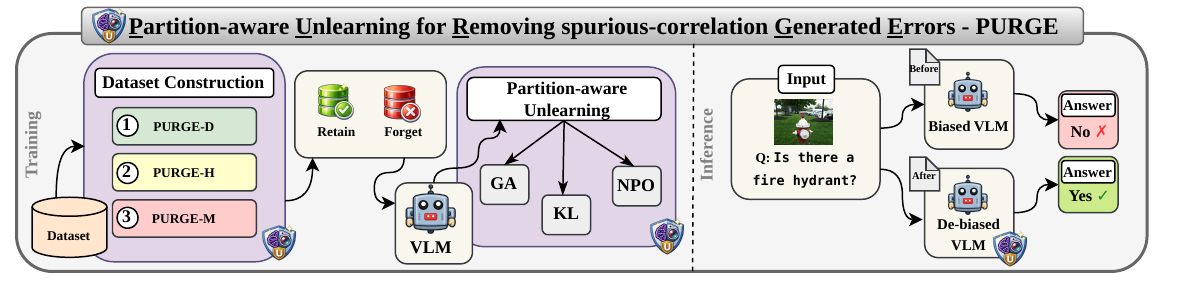}
\vspace{-5mm}
\caption{\small{\algo~framework: Structured data construction followed by partition-aware unlearning}
}

\label{fig:framework}
\end{figure}

In this work, we propose \textbf{\algo~}(\underline{P}artition-aware \underline{U}nlearning for \underline{R}emoving spurious-correlation \underline{G}enerated \underline{E}rrors), a framework for constructing, benchmarking, and mitigating spurious object-background correlations in LVLMs. As illustrated in Fig.~\ref{fig:framework}, \algo~has two main components: -- \one We construct {\em structured retain/forget partitions} using three complementary strategies that expose spurious correlations at different levels: \algo-D, a data-level disentanglement strategy that separates object and background information via masking and inpainting; \algo-H, a hybrid behavior-driven strategy that combines ground-truth annotations, model predictions, and counterfactual inputs to identify shortcut-driven errors; and \algo-M, a model-informed semantic strategy that partitions examples by object presence and contextual cues to isolate background-driven predictions. -- \two We apply {\em partition-aware unlearning}, which uses these structured partitions to suppress background-driven predictions while preserving correct object reasoning, using standard unlearning methods like Gradient Ascent (GA), KL-regularized unlearning (KL), and Negative Preference Optimization (NPO)~\cite{zhang2024negative}. 

\vspace{-4 mm}

We evaluate \algo~across multiple LVLMs, including LLaVA-1.6-7B~\cite{liu2024llavanext}, Qwen3-VL-8B-Instruct~\cite{DBLP:journals/corr/abs-2511-21631}, and Qwen3.5-9B~\cite{qwen3.5}, together with CLIP as a vision-language encoder, on a diverse suite of benchmarks including CHAIR~\cite{rohrbach2018object}, POPE~\cite{li2023evaluating}, Causal-HalBench, MM-SpuBench~\cite{ye2026mm}, AMBER~\cite{wang2023amber}, MMHal~\cite{sun2024aligning}, and Waterbirds. Across these settings, \algo~consistently reduces hallucination and spurious-correlation-driven errors while maintaining or improving overall performance, as shown in Fig.~\ref{fig:chair_intro}. Our contributions are summarized as follows: 


\RA We identify spurious object-background correlations as an important
contributor to hallucination in LVLMs, and introduce structured
object-grounded and context-only retain/forget partitions that enable
joint diagnosis and targeted unlearning of these dependencies.
 \vspace{1 mm}\\ 
\RA We introduce \algo, a unified two-stage framework for constructing, benchmarking, and mitigating spurious-correlation-induced errors through structured retain/forget partitioning and partition-aware unlearning. 
\vspace{-3 mm}
\begin{itemize}[leftmargin=4mm]
\item[{$\bullet$}] We develop three complementary data construction strategies: data-level disentanglement, hybrid behavior-driven partitioning, and model-informed semantic partitioning, which capture spurious correlations across pixel-level evidence, model behavior, and contextual semantics.
\vspace{-1 mm}
\item[{$\bullet$}] We propose partition-aware unlearning as a targeted mechanism for suppressing spurious object-background associations while preserving object-grounded reasoning, using standard unlearning objectives including GA, KL, and NPO.
\vspace{-2 mm}
\end{itemize}
\RA We demonstrate consistent improvements across diverse LVLMs and vision-language models, including LLaVA-1.6-7B, Qwen3-VL-8B-Instruct, Qwen3.5-9B, and CLIP, on multiple hallucination and spurious-correlation benchmarks, reducing hallucination and improving robustness while preserving
competitive overall utility.
\raggedbottom
\vspace{-2 mm}
~\section{Related work}
Spurious correlations have been widely studied in LLMs and LVLMs ~\cite{arjovsky2019invariant,udomcharoenchaikit2022mitigating}, where models may rely on background cues~\cite{hosseini2026spurlens}, language priors~\cite{vo2026vision}, or dataset co-occurrence statistics rather than object relevant evidence ~\cite{sagawa2019distributionally}. In multimodal settings, these shortcuts are especially challenging because 
visual and textual representations are often entangled, amplifying hallucinations and degrading out-of-distribution generalization. Existing mitigation methods can be broadly grouped into training-time,
inference-time, and unlearning-based approaches.

\noindent
\textbf{Training-time debiasing.} Training-time methods aim to reduce spurious correlations by modifying model parameters, representations, or supervision signals. Multimodal contrastive debiasing
approaches separate shortcut cues from semantic alignments by discouraging spurious image-text
associations ~\cite{yang2023mitigating,lu2025mitigating}. Prompt-based methods steer models away from biased directions using designed or reweighted prompts, offering parameter-efficient alternatives to full finetuning~\cite{chuang2023debiasing,jiang2025debiased}.
Region-aware methods identify and suppress spurious image regions, showing that many failures arise from background and contextual cues~\cite{varma2024ravl,yang2025escaping}. Representation-level methods further manipulate the joint embedding space through feature masking, residual correction, or neuron-level intervention
to suppress biased components ~\cite{jung2024unified, seth2023dear, an2026interpretable}. Recent causal approaches also attribute hallucinations to co-occurrence-driven entanglement and seek to disentangle object and context representations within the model~\cite{hu2025causal}. While effective, these methods typically require retraining or additional supervision and may still inherit biases from correlated training distributions.

\noindent
\textbf{Inference-time debiasing.} Inference-time methods mitigate hallucination or bias without updating model parameters. Embedding-level approaches adjust intermediate representations through
projection or input-adaptive transformations to suppress spurious components while preserving 
utility ~\cite{gerych2024bendvlm, lian2026closed}. Decoding-level methods operate directly on output distributions: visual contrastive decoding promotes predictions consistent with visual evidence~\cite{leng2024mitigating}, instruction-based contrastive 
decoding suppresses language-driven hallucinations using perturbed prompts~\cite{wang2024mitigating}, and OPERA penalizes over-reliance on internal priors during generation~\cite{huang2024opera}. Although lightweight and deployable, these methods primarily adjust outputs at inference 
time and do not directly remove spurious associations encoded in model representations. 

\noindent
\textbf{Unlearning-based approaches.} Recent work has explored multimodal unlearning as a mechanism for removing harmful or undesirable associations while preserving overall model utility~\cite{chen2026safety, hen2025safeeraser, li2024single, chakraborty2024can}.
These methods apply targeted forgetting to suppress biased, unsafe, or unwanted dependencies 
encoded in model representations. Subsequent approaches extend unlearning to finer-grained settings,
including concept-level and domain-level removal, through attention guidance, targeted updates, 
or cross-modal interventions ~\cite{kawamura2025approximate,geng2025sauce, bhaila2025cross, huo2025mmunlearner}. However, most existing unlearning methods assume that the information to forget can be specified directly and separated from task-relevant knowledge. In LVLMs, object evidence and contextual cues are often tightly entangled, making it difficult to define what should be forgotten without also damaging useful object-grounded reasoning.

Despite substantial progress, existing methods leave two key gaps. First, they provide limited
mechanisms for constructing controlled partitions that isolate when LVLM predictions are object grounded versus shortcut-driven. Second, they either mitigate shortcuts during training, correct 
outputs at inference time, or perform unlearning without explicitly modeling object-background 
entanglement. PURGE addresses these gaps by combining structured data partitioning with partition aware unlearning: the partitions provide a controlled evaluation protocol for measuring shortcut
reliance and define targeted retain/forget sets to remove spurious object-background associations while preserving object-based reasoning.

\section{\algo:~Structured data partitioning for spurious correlation analysis}
\label{sec:data}
In this section, we present the data construction component of \algo, which builds structured retain and forget partitions for controlled evaluation and mitigation of spurious object-background correlations. We model an input image as $v=(o,z)$, where $o$ denotes object-specific visual evidence and $z$ denotes contextual or background features. Given a target concept $c$, an LVLM with parameters $\theta$ receives a multimodal input $x=(v,q)$, where $q$ is a text query, and predicts $P_{\theta}(c\mid x)$. Our goal is to construct structured retain and forget sets, denoted by $D_r$ and $D_f$, that separate object-grounded evidence from shortcut-inducing contextual cues. We instantiate this idea using MSCOCO~\cite{lin2014microsoft} and Waterbirds~\cite{sagawa2019distributionally}, and we design three complementary partitioning strategies, as discussed next.

\subsection{\algo-D (Data-level disentanglement)} 
\begin{figure}[b]
    \centering
    \includegraphics[width=\textwidth]{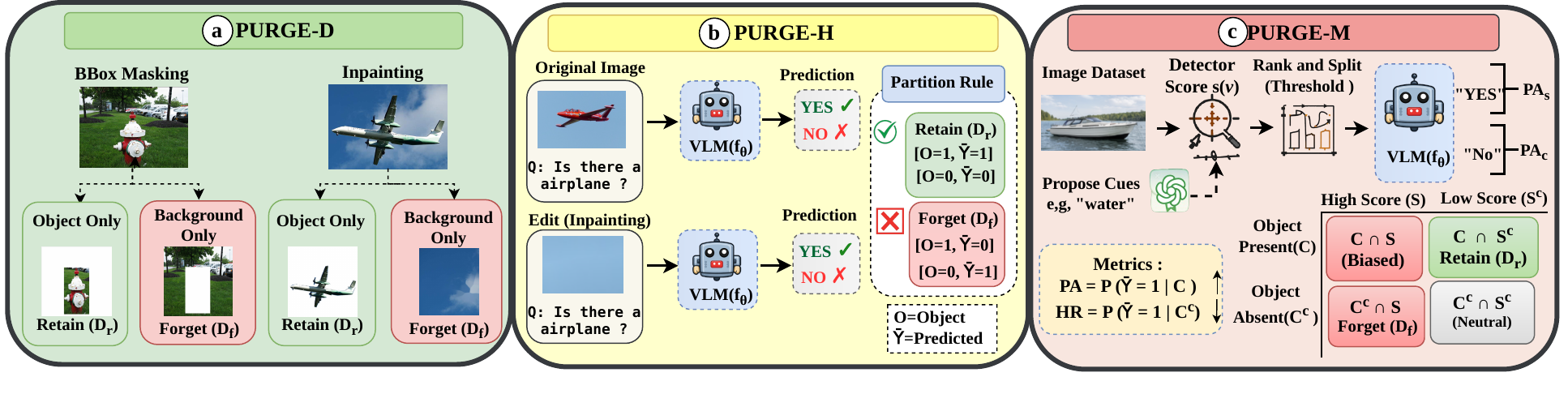}
    \caption{\small Three complementary partitioning strategies in~\algo: 
    {(a) \algo-D} performs data-level disentanglement via masking and inpainting, 
    {(b) \algo-H} applies hybrid behavior-driven partitioning based on prediction correctness, and 
    {(c) \algo-M} performs model-informed semantic disentanglement via cue scoring.}
    \label{fig:strategies}
\end{figure}

As illustrated in Fig.~\ref{fig:strategies}a, \algo-D separates object and background information directly at the image level using masking (or inpainting). Given an image $v$ and an object mask $M(v)$, we construct two complementary views: $v_r = v \odot M(v)$, $v_f = v \odot (1 - M(v)),$ where $v_r$ preserves object-centric information and $v_f$ removes the object while retaining contextual cues. We then partition the dataset into retain and forget sets as:
\vspace{-2 mm}
\begin{tcolorbox}[title= \algo-D]
\vspace{-2 mm}
{\bf Retain set} ${D_r^{(1)}}$: object-focused samples constructed using $v_r$, encouraging predictions based on true object evidence.\\
{\bf Forget set} ${D_f^{(1)}}$: context-only samples constructed using $v_f$, augmented with additional object-absent hard negatives that expose spurious contextual dependencies.
\vspace{-2 mm}
\end{tcolorbox}
We further consider an inpainting-based variant~\cite{wang2018image, Suvorov_2022_WACV}, where removed regions are filled to produce more natural counterfactual images. This variant is reported in App.~\ref{app:IP}. While \algo-D provides a direct pixel-level intervention, it may not fully capture higher-level semantic or model-specific shortcut behavior.


\subsection{\algo-H (Hybrid behavior-driven partitioning)} As illustrated in Fig.~\ref{fig:strategies}b, \algo-H combines object-level supervision with model behavior under original and counterfactual inputs. For each image $v$, we use ground-truth annotations to determine whether the target object is present, denoted by $o\in\{0,1\}$. We then query the LVLM with a binary question, such as ``Is there a $c$ in the image?'', and obtain the prediction $\hat{y}$. To probe reliance on contextual cues, we also construct a counterfactual image $v'$ by removing the target object and query the model again. This yields four cases: 
$$ \underbrace{(o=1,\hat{y}=1)}_{\text{Correct detections}}, ~\underbrace{(o=1,\hat{y}=0)}_{\text{Missed detections}},~\underbrace{(o=0,\hat{y}=1)}_{\text{Hallucinations}},~ \underbrace{(o=0,\hat{y}=0)}_{\text{Correct rejections}},$$ 
using this we partition the dataset as:
\vspace{-2 mm}
\begin{tcolorbox}[title= \algo-H]
\vspace{-2 mm}
{\bf Retain set} ${D_r^{(2)}}$: Reliable prediction cases where the model behaves correctly, including correct detections $(1,1)$ and correct rejections $(0,0)$.\\
{\bf Forget set} ${D_f^{(2)}}$: Failure cases including missed detections $(1,0)$ and hallucinations $(0,1)$, where predictions deviate from the true object presence. 
\vspace{-2 mm}
\end{tcolorbox}
In particular, hallucination cases $(0,1)$ correspond to scenarios in which the model predicts objects purely from contextual cues despite the absence of objects, directly capturing spurious correlations. Compared to purely data-driven approaches, {\algo-H} explicitly captures model-specific biases and failure modes, making it effective for reducing hallucinations. However, imperfect counterfactual generation may introduce noise and partial model dependence.

\subsection{\algo-M (Model-informed semantic partitioning)} 
As illustrated in Fig.~\ref{fig:strategies}c, \algo-M separates object evidence from contextual cues at the semantic level utilizing the learned model. Let $C$ denote the set of images where the target object is present, and let $C^c$ denote its complement. Similarly, let $S$ denote the set of images containing a candidate spurious background cue, and let $S^c$ denote the corresponding low-cue set. Following~\cite{hosseini2026spurlens}, we first use a language model to propose candidate co-occurring cues and then use an open-vocabulary detector to compute per-image cue scores $s(v)$, which are thresholded to define $S$ and $S^c$. This produces four semantic partitions: $ C \cap S,~ C \cap S^c,~ C^c \cap S,~ C^c \cap S^c$ which are then utilized to partition the dataset as:
\vspace{-2 mm}
\begin{tcolorbox}[title = \algo-M]
    \vspace{-2 mm}
{\bf Retain set} ${D_r^{(3)} = C \cap S^c}$: Object-centric samples where the target object is present without strong contextual bias, providing reliable supervision for object-grounded learning.\\
{\bf Forget set} ${D_f^{(3)} = C^c \cap S}$: Hallucination-prone samples where contextual cues are present despite object absence, directly capturing spurious object-background dependencies. 
    \vspace{-2 mm}
\end{tcolorbox}
The remaining partitions are excluded from training: $C\cap S$ contains entangled object-context evidence, while $C^c\cap S^c$ contains trivial negatives with limited disentanglement signal. Both are retained for evaluation. Using $D_r^{(3)}$ and $D_f^{(3)}$, \algo-M encourages the model to suppress context-driven predictions $P(c\mid z)$ while preserving object-based predictions $P(c\mid o)$. Further details of the \algo-M cue-discovery, scoring, selection, and partition-construction pipeline are provided in Appendix~\ref{app:purge_m_construction}.

\section{\algo:~Partition-aware unlearning for spurious correlation mitigation}
We now present the partition-aware unlearning component of \algo, which uses the structured partitions from Sec.~\ref{sec:data} to suppress context-driven predictions while preserving object-grounded reasoning. Ideally, predictions should be grounded in object evidence, so that
$P_{\theta}(c\mid x)$ is primarily determined by $P_{\theta}(c\mid o)$ rather than by contextual cues. However, under spurious object-background correlations, an LVLM may instead rely on the context $z$, leading to high $P_{\theta}(c\mid z)$ even when the target object is absent. This produces hallucinations in cases where $o=0$ but background cues associated with the concept are present. Let $\theta'$ denote the parameters after unlearning. Our goal is to suppress context-driven predictions while preserving object-grounded reasoning:
\begin{equation}
P_{\theta'}(c \mid z) \rightarrow 0, 
\quad \text{while} \quad 
P_{\theta'}(c \mid o) \approx P_{\theta}(c \mid o).
\end{equation}
We formulate this goal as a targeted unlearning problem over the structured partitions constructed in Sec.~\ref{sec:data}. For each partitioning strategy $k\in\{1,2,3\}$, we utilize the retain set $D_r^{(k)}$ containing samples that support object-grounded reasoning, and the forget set $D_f^{(k)}$ containing samples that expose spurious or failure-inducing context dependence. The general objective is
\begin{equation}
    \mathcal{L}^{(k)}
    =
    \mathcal{L}_{\mathrm{retain}}^{(k)}
    +
    \lambda \mathcal{L}_{\mathrm{forget}}^{(k)},
    \label{eq:general}
\end{equation}
where $\mathcal{L}_{\mathrm{retain}}^{(k)}$ preserves useful object-based predictions on $D_r^{(k)}$, $\mathcal{L}_{\mathrm{forget}}^{(k)}$ suppresses spurious predictions on $D_f^{(k)}$, and $\lambda>0$ controls the strength of forgetting.

\noindent
\textbf{Training setup.} We train a LVLM on multimodal inputs $x = (v, q)$, where $v$ denotes the input image and $q$ is a text query (e.g., \textit{``Is there a $c$ in the image?''}). The model predicts $ P_\theta(y \mid v, q), \quad y \in \{\text{yes}, \text{no}\}$. We adopt parameter-efficient fine-tuning $\theta = \{\phi, \theta_{\text{LoRA}}\}$ where $\phi$ denotes the multimodal projector parameters and $\theta_{\text{LoRA}}$ represents the LoRA parameters of the language model. The vision encoder and backbone of LVLM remain frozen. Let $(v_r, q_r, y_r) \sim D_r^{(k)}$ and $(v_f, q_f, y_f) \sim D_f^{(k)}$. We instantiate Eq.~\eqref{eq:general} with three unlearning objectives,  Gradient Ascent (GA), KL-regularized unlearning (KL), and Negative Preference Optimization (NPO)~\citep{zhang2024negative}, each offering a different trade-off between forgetting strength and stability: \vspace{1 mm}\\
-- \one \textbf{GA} preserves correct predictions on the retain set while reducing the likelihood of spurious labels on the forget set:
\begin{align}
\mathcal{L}_{\text{GA}}^{(k)} &\coloneq
    -\,\mathbb{E}_{D_r^{(k)}} \log P_\theta(y_r \mid v_r, q_r)
    +\,\lambda\,\mathbb{E}_{D_f^{(k)}} \log P_\theta(y_f \mid v_f, q_f).
    \label{eq:ga}
\end{align}
Since the objective is minimized, the second term decreases the probability assigned to the forget labels. GA is simple and aggressive, but it does not explicitly constrain the updated model to remain close to the original model, which can cause instability or unintended degradation. \vspace{1 mm}\\
-- \two \textbf{KL} regularized unlearning suppresses forget-set predictions while keeping the updated model close to a reference model $P_{\theta_0}$ on retain samples to improve stability of GA:
\begin{align}
\mathcal{L}_{\text{KL}}^{(k)} &= 
    \lambda\,\mathbb{E}_{D_f^{(k)}} \log P_\theta(y_f \mid v_f, q_f)
    + \beta\,\mathbb{E}_{D_r^{(k)}} \mathrm{KL}\!\left(P_\theta(\cdot|v_r,q_r) \,\|\, 
    P_{\theta_0}(\cdot|v_r,q_r)\right).
    \label{eq:kl}
\end{align}
Here, $\theta_0$ denotes the original model before unlearning. The KL term reduces drift on retain samples, although strong regularization can weaken forgetting when retain and forget objectives conflict. \vspace{1 mm}\\
-- \three \textbf{NPO} penalizes spurious predictions relative to the reference model rather than only in absolute likelihood. This discourages the updated model from assigning high confidence to forget-set labels while maintaining retain-set performance:
\begin{align}
\mathcal{L}_{\text{NPO}}^{(k)} &= 
    -\,\mathbb{E}_{D_r^{(k)}} \log P_\theta(y_r \mid v_r, q_r)
    + \frac{2\lambda}{\beta}\,\mathbb{E}_{D_f^{(k)}} 
    \log\!\left(1 + \left(\frac{P_\theta(y_f|v_f,q_f)}
    {P_{\theta_0}(y_f|v_f,q_f)}\right)^{\!\beta}\right).
    \label{eq:npo}
\end{align}
The hyperparameter $\beta$ controls the sharpness of the relative penalty. Compared with GA, NPO provides a more targeted forgetting signal by focusing on forget labels whose confidence remains high relative to the reference model. Together, these objectives span different forgetting behaviors, from aggressive forgetting with GA to more conservative regularization with KL and relative preference-based suppression with NPO. The three partitioning strategies in Sec.~\ref{sec:data} instantiate the retain and forget sets as follows:
\vspace{-1.5 mm}
\begin{tcolorbox}[title = Partition-aware unlearning: \algo]
\vspace{-2 mm}
    \textbf{\algo-D}: Retain samples are object-only views and forget samples are background-only views, with $y_r {=} y_f {=}  \text{yes}$, so the forget term suppresses spurious background confidence. Since $D_f^{(1)}$ contains only background-only images without a natural negative label, an auxiliary set $D_f^-$ provides explicit \textit{no} supervision via an additional term {$\mathcal{L}_{\text{neg}}^{(1)}=-\mathbb{E}_{D_f^-} \log P_\theta(\text{no} \mid v, q)$}, giving $\mathcal{L}^{(1)} = \mathcal{L}_{\text{retain}}^{(1)} + \lambda\, \mathcal{L}_{\text{forget}}^{(1)}+\gamma\mathcal{L}_{\text{neg}}^{(1)}$ with $\gamma > 0$. \vspace{1 mm}\\
  \textbf{\algo-H}: $y_r$ corresponds to the model's correct predictions and $y_f$ corresponds to the model's spurious predictions covering both object-present and object-absent cases. Note that the $(0,0)$ cases in $D_r^{(2)}$ already provide negative supervision,  as object-absent samples with \textit{no} labels are naturally included in the retain set. \vspace{1 mm}\\
  \textbf{\algo-M}: $y_r {=} y_f {=} \text{yes}$, with retain drawn from $C \cap S^c$ and forget from $C^c \cap S$. Since all forget samples are object-absent, the forget objective itself directly penalizes false-positive predictions on object-absent images, making explicit negative supervision redundant.
\vspace{-2 mm}
\end{tcolorbox}

\section{Experimental evaluation} 
\label{sec:Exp}
We evaluate \algo~along three dimensions: 
-- \one Whether partition-aware unlearning reduces hallucination and spurious-correlation-driven errors, 
-- \two Whether the improvements generalize across LVLM architectures and benchmarks, and 
-- \three How different partitioning strategies and trainable components affect performance.

\noindent\textbf{Models.} We evaluate the \algo~framework on LLaVA-1.6-7B~\cite{liu2024llavanext}, Qwen3-VL-8B-Instruct~\cite{DBLP:journals/corr/abs-2511-21631}, and Qwen3.5-9B~\cite{qwen3.5}. LLaVA follows a modular design with a CLIP-based~\cite{radford2021learning} vision encoder and language model, while Qwen3-VL adopts a unified multimodal architecture with strong vision-language alignment. For Qwen3.5-9B, vision--language fusion is tightly integrated within the multimodal architecture; therefore, we do not train a separate multimodal projector. Adaptation is performed through LoRA-based updates while the integrated multimodal fusion architecture is preserved.

\noindent\textbf{Baselines.} We compare against the corresponding vanilla models without unlearning. To isolate the role of partition-aware data construction, we evaluate standard unlearning objectives, including GA, KL, and NPO, applied to the constructed partitions. In addition, we compare with prior multimodal debiasing methods such as RAVL~\cite{varma2024ravl}, and with CLIP-based debiasing on Waterbirds~\cite{yang2023mitigating}.

\noindent\textbf{Benchmarks and metrics.} We evaluate on a diverse set of benchmarks covering object hallucination, object existence, causal consistency, and robustness to spurious correlations, including CHAIR~\cite{rohrbach2018object}, AMBER~\cite{wang2023amber}, POPE~\cite{li2023evaluating}, MMHal~\cite{sun2024aligning}, Causal-HalBench~\cite{xu2026causal}, MM-SpuBench~\cite{ye2026mm}, and Waterbirds~\cite{sagawa2019distributionally}. Details of each benchmark are provided in App.~\ref{app:EB}. In all tables, green shading indicates the best result and red shading indicates the second-best result.

\vspace{-2mm}

\noindent
\begin{minipage}[t]{0.53\textwidth}
\vspace{0pt}

\textbf{Spurious-correlation robustness on Waterbirds with CLIP.}
Table~\ref{tab:clip_waterbirds} evaluates CLIP-RN50 on
Waterbirds~\cite{sagawa2019distributionally}. Starting from Waterbirds,
we construct PURGE-specific retain and forget partitions based on object
and background cues and use these partitions for partition-aware
unlearning. \algo-M-NPO achieves the highest WGA among the evaluated
methods ($79.6$) while maintaining a competitive average accuracy of
$90.9$, demonstrating reduced reliance on spurious background
correlations while preserving overall performance.

\end{minipage}
\hspace{0.003\textwidth}
\begin{minipage}[t]{0.46\textwidth}
\vspace{0pt}
\centering

\captionsetup{
    justification=centering,
    skip=4pt
}

\captionof{table}{\small CLIP on Waterbirds.}
\label{tab:clip_waterbirds}

\scriptsize
\setlength{\tabcolsep}{3.6pt}
\renewcommand{\arraystretch}{0.65}

\setlength{\aboverulesep}{1pt}
\setlength{\belowrulesep}{1pt}
\setlength{\abovetopsep}{0pt}
\setlength{\belowbottomsep}{0pt}

\begin{tabular}{@{}lcc@{}}
\toprule
\textbf{Method} &
\textbf{Avg.\ Acc.\ $\uparrow$} &
\textbf{WGA $\uparrow$} \\
\midrule

\multicolumn{3}{@{}l}{\textbf{CLIP baselines}} \\
Pre-trained CLIP & 90.8 & 44.9 \\
Fine-tuned CLIP  & 81.3 & 77.1 \\

\midrule
\multicolumn{3}{@{}l}{\textbf{Existing methods}} \\
ERM & \best{93.5} & 54.4 \\
Group DRO~\cite{sagawa2019distributionally} & 83.3 & 73.7 \\
MSC~\cite{yang2023mitigating} & 84.7 & 77.5 \\
MSC[variant]~\cite{yang2023mitigating} & 83.2 & 77.5 \\

\midrule
\multicolumn{3}{@{}l}{\textbf{Partition-aware unlearning}} \\
\algo-D-NPO & 83.5 & 75.1 \\
\algo-H-NPO & 86.3 & \second{79.3} \\
\algo-M-NPO & \second{90.9} & \best{79.6} \\

\bottomrule
\end{tabular}

\end{minipage}
\par
\noindent
\noindent\textbf{Comparison methods.}
Pre-trained CLIP is the original CLIP-RN50, while Fine-tuned CLIP adapts it to our Waterbirds-derived data by updating the visual encoder and classification head with the text encoder frozen. ERM uses the same CLIP-RN50 initialization, Group DRO follows Sagawa et al.~\cite{sagawa2019distributionally}, and MSC/MSC[variant] are the two multimodal spurious-correlation mitigation settings from Yang et al.~\cite{yang2023mitigating}.

\raggedbottom

\begin{wraptable}[14]{r}{0.54\textwidth}
    \vspace{0pt}
    \centering
    \captionsetup{justification=centering}
    \caption{\small Robustness across models.}
    \label{tab:robust_spurious}
    \vspace{-2pt}

      \tiny
    \setlength{\tabcolsep}{2pt}
    \renewcommand{\arraystretch}{0.70}

    \resizebox{\linewidth}{!}{%
    \begin{tabular}{@{}llcc@{}}
    \toprule
    \textbf{Model} & \textbf{Method} & \textbf{Overall $\uparrow$} & \textbf{WGA $\uparrow$} \\
    \midrule

    Prior work & RAVL~\cite{varma2024ravl} & 70.2 & 40.8 \\
    \midrule

    \multirow{3}{*}{LLaVA-1.6-7B}
        & Baseline       & 54.4 & 32.0 \\
        & \algo-M-KL     & 72.1 & 46.4 \\
        & \algo-M-NPO    & \second{77.5} & \second{60.1} \\
    \midrule

    \multirow{3}{*}{\shortstack[l]{Qwen3-VL\\8B-Instruct}}
        & Baseline       & 62.7 & 46.3 \\
        & \algo-M-KL     & 76.8 & 54.2 \\
        & \algo-M-NPO    & \best{82.9} & \best{62.8} \\
    \midrule

    \multirow{3}{*}{Qwen3.5-9B}
        & Baseline       & 59.8 & 47.5 \\
        & \algo-M-KL     & 74.1 & 51.4 \\
        & \algo-M-NPO    & 77.2 & 59.2 \\
    \bottomrule
    \end{tabular}%
    }

    \vspace{-2mm}
\end{wraptable}

\noindent\textbf{Robustness across models.}
Table~\ref{tab:robust_spurious} evaluates \algo-M on the held-out test
split of our constructed object--context dataset, using average
classification accuracy (Overall) and worst-group accuracy (WGA) over
the four object--context groups. RAVL score averages CLIP-RN50/RN101. Across all evaluated LVLMs, \algo-M consistently improves both overall accuracy and WGA over the
corresponding baselines, with especially large gains in WGA. These
results suggest that the proposed partitions expose shortcut-driven
behavior in a way that enables effective and architecture-agnostic
mitigation.

\par

\begin{wraptable}[15]{r}{0.47\textwidth}
    \vspace{-4mm}
    \centering
    \captionsetup{justification=centering}
    \caption{\small Evaluation on Causal-HalBench.}
    \label{tab:causal}
    \vspace{-2pt}

    \scriptsize
    \setlength{\tabcolsep}{2.4pt}
    \renewcommand{\arraystretch}{0.94}

    \resizebox{\linewidth}{!}{%
    \begin{tabular}{lcccccccc}
    \toprule
    \textbf{Method}
    & \textbf{$Q_c$}
    & \textbf{$Q_a$}
    & \textbf{$Q_c^\prime$}
    & \textbf{$Q_a^\prime$}
    & \textbf{Acc. $\uparrow$}
    & \textbf{CAC $\downarrow$}
    & \textbf{AAC $\downarrow$}
    & \textbf{CHR $\downarrow$} \\
    \midrule

    \multicolumn{9}{c}{\textbf{LLaVA-1.6-7B}} \\
    \midrule

    Baseline
    & 80.2 & 81.1 & 75.7 & 82.9 & 85.6 & 4.5 & 1.8 & 12.9 \\

    \algo-D-NPO
    & 85.1 & 85.8 & 81.3 & 86.9 & 91.1 & 3.8 & 1.1 & 10.5 \\

    \algo-H-NPO
    & 88.3 & 87.2 & 85.4 & 88.1 & 93.1 & 2.9 & 0.9 & 6.7 \\

    \algo-M-NPO
    & \second{91.3}
    & \second{93.3}
    & 89.6
    & \second{93.6}
    & \best{94.6}
    & 1.7
    & 0.3
    & 3.2 \\

    \midrule
    \multicolumn{9}{c}{\textbf{Qwen3-VL-8B-Instruct}} \\
    \midrule

    Baseline
    & 81.8 & 83.2 & 78.3 & 84.4 & 84.7 & 3.5 & 1.2 & 7.4 \\

    \algo-D-NPO
    & 86.9 & 87.7 & 84.0 & 88.6 & 90.0 & 2.9 & 0.9 & 3.2 \\

    \algo-H-NPO
    & 91.1
    & 93.3
    & \second{89.7}
    & 93.5
    & 93.6
    & \second{1.4}
    & \second{0.2}
    & \second{1.9} \\

    \algo-M-NPO
    & \best{91.7}
    & \best{94.3}
    & \best{91.0}
    & \best{94.2}
    & \second{94.1}
    & \best{0.7}
    & \best{0.1}
    & \best{1.2} \\

    \midrule
    \multicolumn{9}{c}{\textbf{Qwen3.5-9B}} \\
    \midrule

    Baseline
    & 78.2 & 76.6 & 73.1 & 78.3 & 82.3 & 5.1 & 1.7 & 11.6 \\

    \algo-D-NPO
    & 80.1 & 78.5 & 75.3 & 79.7 & 84.8 & 4.8 & 1.2 & 9.7 \\

    \algo-H-NPO
    & 82.3 & 81.4 & 78.3 & 82.4 & 86.4 & 4.0 & 1.0 & 6.8 \\

    \algo-M-NPO
    & 87.5 & 84.8 & 84.3 & 85.4 & 90.4 & 3.2 & 0.6 & 4.3 \\

    \bottomrule
    \end{tabular}%
    }

    \vspace{-2mm}
\end{wraptable}

\noindent\textbf{Spurious-correlation robustness and causal consistency.}
Table~\ref{tab:causal} further evaluates causal consistency on Causal-HalBench.
The metrics $Q_c$ and $Q_a$ denote contextual-object and absent-object
questions, respectively. CAC and AAC measure the corresponding accuracy
changes between the original and counterfactual images. CHR denotes the
counterfactual-object hallucination rate. Lower CAC, AAC, and CHR indicate
better causal robustness. Across models, \algo~improves $Q_c$ and $Q_a$,
reduces CAC and AAC, and lowers CHR. These results show that \algo~not only
improves accuracy but also makes predictions more stable under counterfactual
changes. Detailed group-wise MM-SpuBench results are reported in App.~\ref{app:MM}.

\par
\begin{figure}[H]
    \centering
    \begin{subfigure}[t]{0.32\textwidth}
        \centering
        \includegraphics[width=\linewidth]{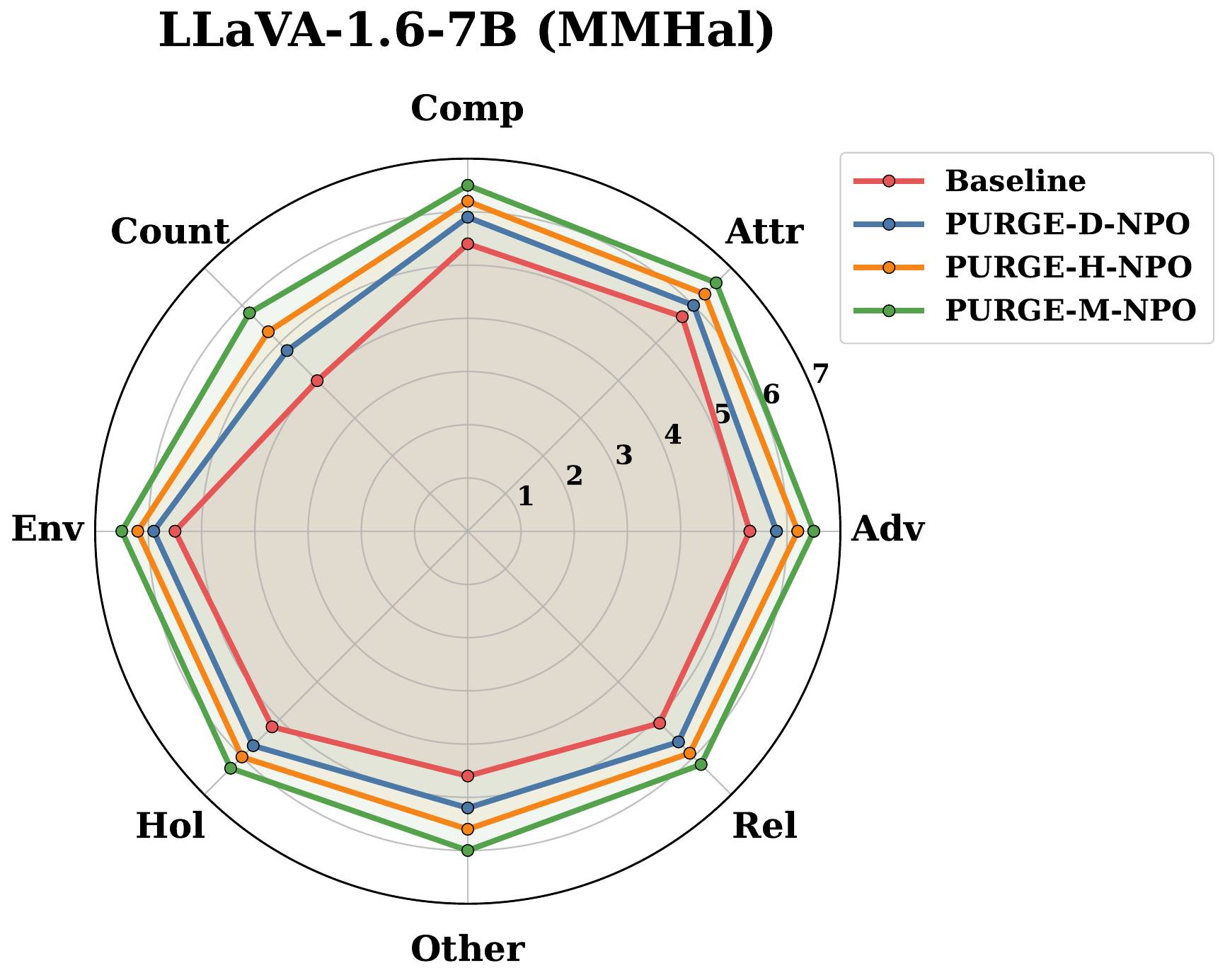}
        \caption{}
        \label{fig:mmhal_llava}
    \end{subfigure}\hfill
    \begin{subfigure}[t]{0.32\textwidth}
        \centering
        \includegraphics[width=\linewidth]{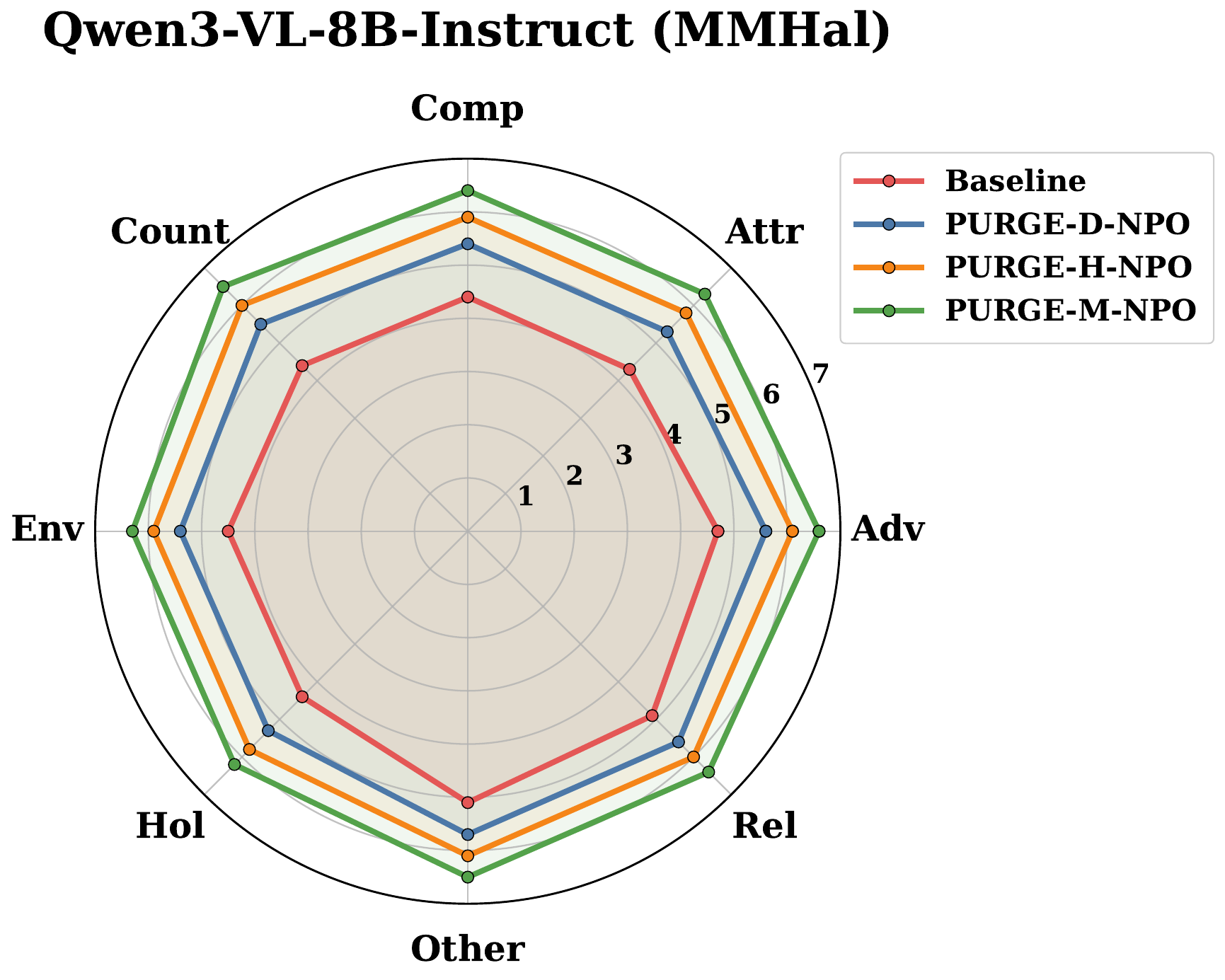}
        \caption{}
        \label{fig:mmhal_qwen3vl}
    \end{subfigure}\hfill
    \begin{subfigure}[t]{0.32\textwidth}
        \centering
        \includegraphics[width=\linewidth]{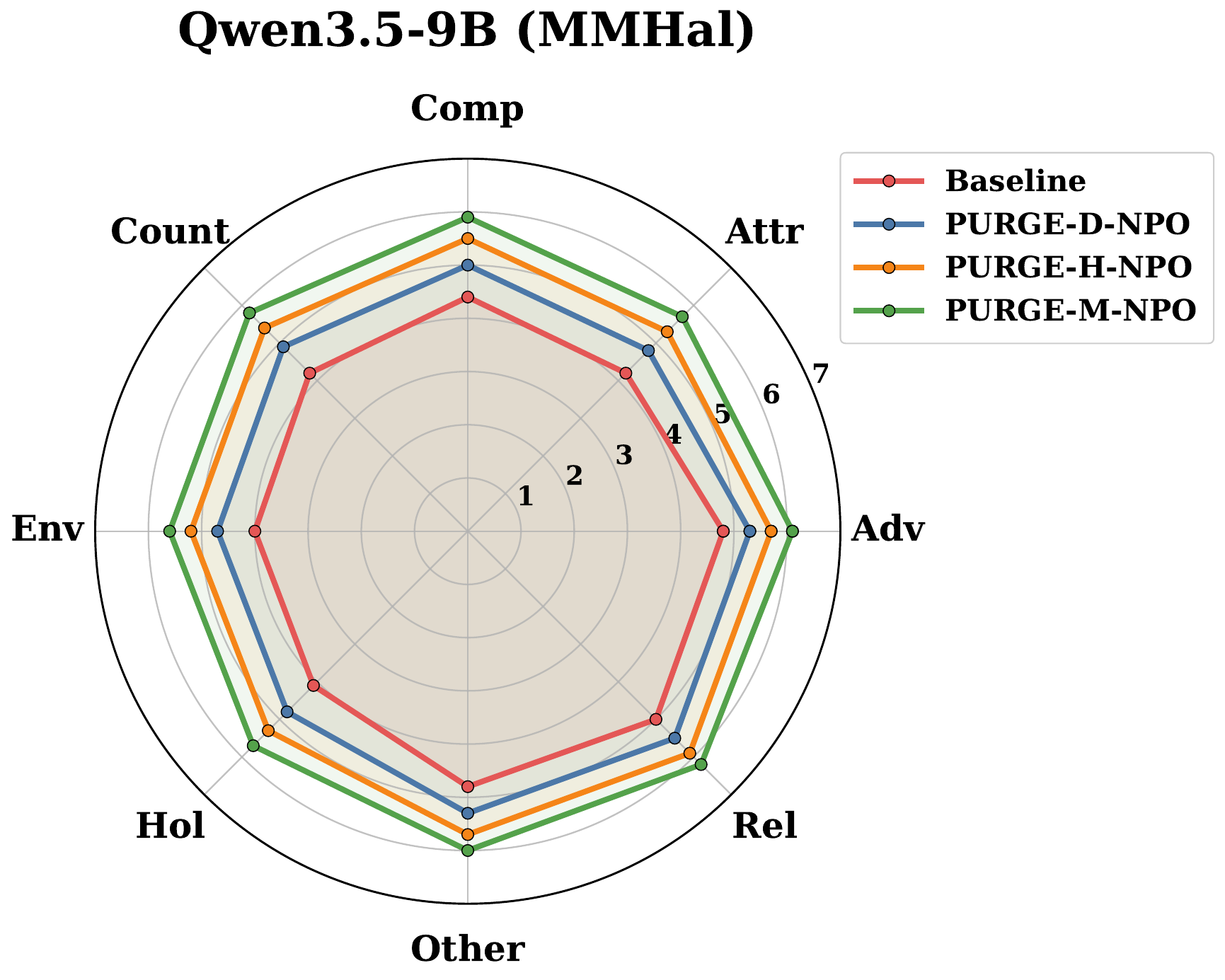}
        \caption{}
        \label{fig:mmhal_qwen35}
    \end{subfigure}
    \caption{\small Category-wise MMHal~\cite{sun2024aligning} evaluation across different models.}
   
    \label{fig:mmhal_spider_all}
\end{figure}

\noindent\textbf{Category-wise MMHal analysis.}
Fig.~\ref{fig:mmhal_spider_all} reports category-wise results on MMHal~[29].
Hallucination and shortcut reliance are more pronounced in categories
such as counting and adversarial examples, where models must distinguish
object evidence from misleading context. Across models, \algo~variants
consistently improve MMHal scores across categories, with \algo-M
achieving the largest overall gains. Since higher MMHal scores indicate
better performance, these improvements demonstrate stronger object
grounding and robustness across diverse reasoning tasks. We further
evaluate the inpainting-based \algo-D variant on MMHal, showing
consistent category-wise improvements; detailed results are provided
in Appendix~\ref{app:IP}.

\vspace{-3mm}

\vspace{-4mm}
\noindent
\textbf{Object-existence evaluation on POPE.} Table~\ref{tab:pope_random} reports results on the POPE Random split, which evaluates object-presence prediction. The vanilla baselines achieve lower accuracy and F1, indicating that models often fail to reliably determine whether an object is present. All \algo~variants improve performance, suggesting stronger object grounding and reduced hallucination. \algo-M-NPO is generally the strongest variant across models, while \algo-D-NPO and \algo-H-NPO achieve the best performance in several model--split combinations.

 \par

\raggedbottom

\begin{table}[H]
\centering
\footnotesize

\caption{\small POPE~\cite{li2023evaluating} results on the Random split.}
\label{tab:pope_random}

\resizebox{\textwidth}{!}{
\begin{tabular}{llcccccccccccc}
\toprule
\multirow{2}{*}{\textbf{Method}}
& \multirow{2}{*}{\textbf{Obj.}}
& \multicolumn{4}{c}{\textbf{LLaVA-1.6-7B}}
& \multicolumn{4}{c}{\textbf{Qwen3-VL-8B-Instruct}}
& \multicolumn{4}{c}{\textbf{Qwen3.5-9B}} \\
\cmidrule(lr){3-6}
\cmidrule(lr){7-10}
\cmidrule(lr){11-14}

& 
& \textbf{Acc} & \textbf{Rec} & \textbf{F1} & \textbf{Yes}
& \textbf{Acc} & \textbf{Rec} & \textbf{F1} & \textbf{Yes}
& \textbf{Acc} & \textbf{Rec} & \textbf{F1} & \textbf{Yes} \\
\midrule

Baseline & --
& 67.2 & 54.5 & 62.4 & 37.3
& 79.1 & 80.7 & 79.4 & 51.6
& 79.9 & 78.8 & 79.7 & 48.9 \\

\cmidrule(lr){1-14}

\multirow{3}{*}{\algo-D}
& GA
& 87.5 & 88.0 & 87.6 & 50.5
& 90.1 & 89.2 & 90.0 & 49.1
& 88.2 & 86.1 & 88.0 & 47.9 \\

& KL
& 92.5 & 89.5 & 92.3 & 47.0
& 97.2 & 94.4 & 97.1 & 47.2
& 90.1 & 90.1 & 90.1 & 50.0 \\

& NPO
& 93.7 & 89.9 & 93.5 & 46.2
& 91.1 & 92.8 & 91.2 & 51.7
& \best{92.3} & 91.6 & \best{92.2} & 49.3 \\

\cmidrule(lr){1-14}

\multirow{3}{*}{\algo-H}
& GA
& 89.5 & 88.5 & 89.4 & 49.0
& 88.7 & 89.1 & 88.7 & 50.4
& 86.2 & 85.4 & 86.1 & 49.2 \\

& KL
& 90.3 & 89.6 & 90.2 & 49.3
& 90.5 & 91.1 & 90.6 & 50.6
& 89.4 & 88.3 & 89.3 & 48.9 \\

& NPO
& \second{97.3} & \second{95.1} & \second{97.2} & 47.8
& 98.1 & 96.2 & 98.1 & 48.1
& \second{92.1} & \second{91.7} & \second{92.1} & 49.6 \\

\cmidrule(lr){1-14}

\multirow{3}{*}{\algo-M}
& GA
& 91.6 & 92.1 & 91.6 & 50.5
& 92.3 & 93.0 & 92.4 & 50.7
& 88.1 & 88.3 & 88.1 & 50.2 \\

& KL
& 92.2 & 92.4 & 92.2 & 50.2
& \second{98.2} & \second{96.4} & \second{98.2} & 48.2
& 89.8 & 91.2 & 89.9 & 51.4 \\

& NPO
& \best{98.1} & \best{96.2} & \best{98.1} & 48.1
& \best{99.6} & \best{99.2} & \best{99.6} & 49.6
& 91.6 & \best{92.1} & 91.6 & 50.5 \\

\bottomrule
\end{tabular}
}
\end{table}

\vspace{-4 mm}

\textbf{Ablation study.} We compare LoRA-only updates, applied to the language model attention and MLP layers, with joint LoRA+projector training. As shown in Tables~\ref{tab:ab_MM} and~\ref{tab:ab_CH}, LoRA-only updates already improve MM-SpuBench accuracy and reduce hallucination on CHAIR. Updating the multimodal projector in addition to LoRA further improves performance, especially on CHAIR, indicating that adapting the vision-language interface is important for removing object-background shortcuts. For Qwen3.5-9B, we keep the vision encoder and multimodal fusion module fixed because they are tightly integrated, and apply LoRA only to the language model. Additional CHAIR results and the inpainting-based \algo-D variant are reported in App.~\ref{app:chair} and App.~\ref{app:IP}.

\vspace{-4 mm}

\begin{table}[H]
\centering
\small
\setlength{\tabcolsep}{4pt}
\renewcommand{\arraystretch}{1}

\begin{minipage}[t]{0.49\textwidth}
\centering
\captionsetup{type=table, justification=centering}
\caption{ \small Accuracy (\%) on MM-SpuBench.}
\label{tab:ab_MM}
\vspace{2pt}

\resizebox{\linewidth}{!}{
\begin{tabular}{llccc}
\toprule
\textbf{Method} & \textbf{Setting} 
& \textbf{LLaVA-1.6-7B} 
& \makecell{\textbf{Qwen3-VL-}\\\textbf{8B-Instruct}} 
& \textbf{Qwen3.5-9B} \\
\midrule
Baseline & -- & 59.7 & 73.8 & 60.1 \\
\midrule
\multirow{2}{*}{\algo-D-NPO}
& LLM only  & 61.4 & 74.7 & 61.2 \\
& LLM+Proj. & 63.8 & 78.3 & -- \\
\midrule
\multirow{2}{*}{\algo-H-NPO}
& LLM only  & 63.1 & 78.8 & 61.8 \\
& LLM+Proj. & 66.6 & 82.4 & -- \\
\midrule
\multirow{2}{*}{\algo-M-NPO}
& LLM only  & 64.2 & \second{83.7} & 64.3 \\
& LLM+Proj. & 71.9 & \best{88.7} & -- \\
\bottomrule
\end{tabular}}
\end{minipage}
\hfill
\begin{minipage}[t]{0.49\textwidth}
\centering
\captionsetup{type=table, justification=centering}
\caption{\small Ablation on CHAIR (lower is better).}
\label{tab:ab_CH}
\vspace{2pt}
\setlength{\tabcolsep}{1.7pt}
\renewcommand{\arraystretch}{0.98}

\resizebox{\linewidth}{!}{
\begin{tabular}{llcccccc}
\toprule
\textbf{Method} & \textbf{Setting}
& \multicolumn{2}{c}{\textbf{LLaVA-1.6-7B}}
& \multicolumn{2}{c}{\makecell{\textbf{Qwen3-VL-}\\\textbf{8B-Instruct}}}
& \multicolumn{2}{c}{\textbf{Qwen3.5-9B}} \\
\cmidrule(lr){3-4} \cmidrule(lr){5-6} \cmidrule(lr){7-8}
& 
& \textbf{CHAIR$_s$} & \textbf{CHAIR$_i$}
& \textbf{CHAIR$_s$} & \textbf{CHAIR$_i$}
& \textbf{CHAIR$_s$} & \textbf{CHAIR$_i$} \\
\midrule
Baseline & -- 
& 41.2 & 23.6 & 32.4 & 19.3 & 36.2 & 21.2 \\
\midrule
\multirow{2}{*}{\algo-D-NPO}
& LLM only 
& 38.6 & 18.3 & 24.6 & 17.8 & 24.8 & 18.9 \\
& LLM+Proj. 
& 20.3 & 12.7 & 17.4 & 11.9 & -- & -- \\
\midrule
\multirow{2}{*}{\algo-H-NPO}
& LLM only 
& 30.1 & 17.2 & 22.6 & 16.8 & 23.3 & 17.0 \\
& LLM+Proj. 
& 18.3 & 9.3 & 11.6 & 7.4 & -- & -- \\
\midrule
\multirow{2}{*}{\algo-M-NPO}
& LLM only 
& 29.1 & 15.5 & 20.2 & 13.7 & 20.7 & 14.2 \\
& LLM+Proj. 
& \second{11.3} & \second{6.6} & \best{9.6} & \best{4.3} & -- & -- \\
\bottomrule
\end{tabular}}
\end{minipage}
\end{table}

\vspace{-4 mm}

\raggedbottom

\begin{figure}[H]
\centering
\includegraphics[width=\textwidth]{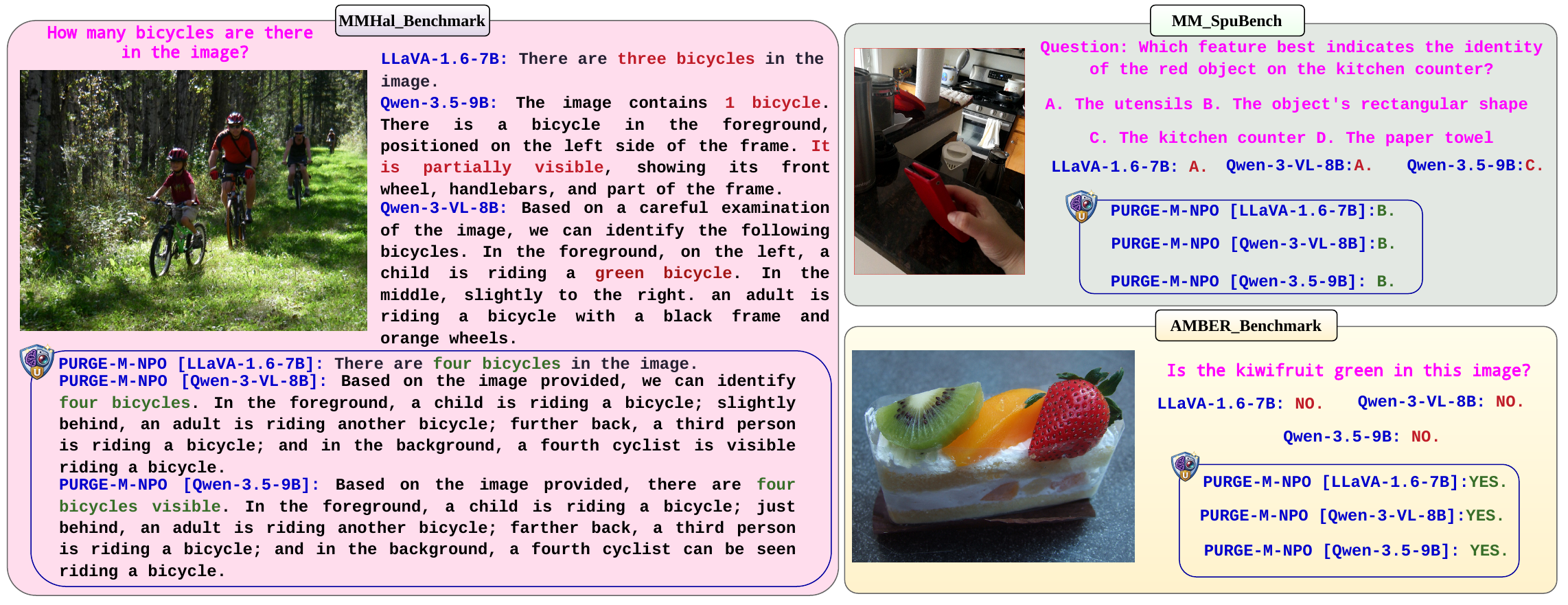}

\caption{\small
{Evaluation across hallucination and spurious correlation benchmarks.}
\textbf{Left:} {MMHal-Bench}, where baseline models undercount objects.
\textbf{Top-right:} {MM-SpuBench} example showing reliance on spurious features.
\textbf{Bottom-right:} {AMBER} example illustrating reduced hallucination
in a binary attribute-grounding query.
}
\label{fig:q}
\end{figure}

\textbf{Qualitative analysis.} Fig.~\ref{fig:q} provides qualitative examples illustrating how spurious correlations contribute to hallucination. Baseline models often rely on misleading contextual cues, leading to incorrect object predictions or counting errors. On MMHal-Bench, this appears as undercounting or overcounting object instances; on MM-SpuBench, predictions are influenced by spurious visual features; and  on AMBER, models produce ungrounded answers in binary attribute-grounding queries. In contrast, \algo~produces responses that are more consistent with object-relevant evidence, improving counting, feature attribution, and object-presence judgments. Quantitative AMBER results are provided in App.~\ref{app:amber}.
\vspace{-4 mm}

\textbf{Controlled Object-Absent Analysis.} To further characterize the behavioral mechanism targeted by PURGE, we
perform a controlled evaluation in which the target object is absent
while contextual cues associated with that object remain available.
Under this setting, predicting the absent object provides a direct
behavioral measure of reliance on contextual shortcuts.
\vspace{-4mm}
\begin{table}[H]
\centering
\caption{\small \textbf{Hallucination rate (\%) on controlled object-absent
examples.} Lower is better.}
\label{tab:object_absent}
\resizebox{0.82\linewidth}{!}{
\begin{tabular}{ccrrrr}
\toprule
Test Set & Image Type &
Base & \algo-D-NPO & \algo-H-NPO & \algo-M-NPO \\
\midrule

PURGE-D &
Background-only &
61.2 & \best{12.3} & \second{18.1} & 21.4 \\

PURGE-H &
Inpainted object-absent &
64.8 & \best{15.8} & \second{19.2} & 22.5 \\

PURGE-M &
Cue-only object-absent &
56.0 & \best{14.9} & \second{17.3} & 19.8 \\
\bottomrule
\end{tabular}}
\end{table}
\vspace{-4mm}
The baseline produces context-driven false-positive predictions in
56--65\% of these controlled object-absent cases. All PURGE variants
substantially reduce this behavior, lowering the hallucination rate to approximately 12--23\%. These findings provide behavioral evidence that PURGE reduces reliance on contextual shortcuts when object-specific evidence is absent. We do not interpret this experiment as establishing a unique causal origin for the learned associations. Rather, biased object--context co-occurrence provides a statistical signal during training, while visual--semantic representation learning determines whether this regularity becomes encoded as a predictive shortcut. PURGE targets such shortcuts once they become behaviorally observable.

\textbf{Prediction Transition and Regression Analysis.} While PURGE is designed to suppress spurious object--background
correlations, contextual information can occasionally provide useful
inductive bias for object recognition. We therefore quantify whether
PURGE introduces new errors by comparing the baseline LLaVA-1.6-7B
model and the corresponding PURGE-M-NPO model. Each sample is categorized as
Correct$\rightarrow$Correct ($C\rightarrow C$),
Incorrect$\rightarrow$Correct ($I\rightarrow C$),
Correct$\rightarrow$Incorrect ($C\rightarrow I$), or
Incorrect$\rightarrow$Incorrect ($I\rightarrow I$).
\vspace{-4mm}
\begin{table}[H]
\centering
\caption{\small Sample-wise prediction transition analysis between
the baseline LLaVA-1.6-7B and PURGE-M-NPO.
Regression denotes a correct baseline prediction that becomes
incorrect after PURGE.}
\label{tab:transition}

\small
\setlength{\tabcolsep}{2 pt}
\renewcommand{\arraystretch}{1}

\resizebox{\linewidth}{!}{%
\begin{tabular}{lcccccccc}
\toprule
\textbf{Benchmark} &
\textbf{\# Samples} &
\textbf{Base Acc.} &
\textbf{PURGE Acc.} &
\textbf{$C\!\rightarrow\!C$} &
\textbf{$I\!\rightarrow\!C$} &
\textbf{$C\!\rightarrow\!I$} &
\textbf{$I\!\rightarrow\!I$} &
\textbf{Regression $\downarrow$} \\
&
&
\textbf{(\%)} &
\textbf{(\%)} &
&
&
&
&
\textbf{(\%)} \\
\midrule

MM-SpuBench
& 2400
& 59.7
& \textbf{78.3}
& 1415
& 464
& 18
& 503
& 0.75 \\

POPE (4K subset)
& 4000
& 72.4
& \textbf{95.6}
& 2,884
& 940
& 12
& 164
& \textbf{0.30} \\

AMBER (4K subset)
& 4000
& 48.6
& \textbf{81.8}
& 1930
& 1342
& 14
& 714
& 0.35 \\

\bottomrule
\end{tabular}%
}
\end{table}
\vspace{-4mm}

For POPE and AMBER, we conduct the sample-wise transition analysis on 4,000 examples, evaluating the same examples with the baseline LLaVA-1.6-7B and with PURGE-M-NPO. Accordingly, the Base Subset Acc. and PURGE Subset Acc. columns in Table~\ref{tab:transition} report the accuracies of the baseline LLaVA-1.6-7B and PURGE-M-NPO, respectively, on these transition-analysis examples. For AMBER, the baseline subset accuracy is obtained directly from the transition counts as $(1930+14)/4000=48.6\%$. Detailed results on the full AMBER benchmark are reported separately in Table~\ref{tab:amber_main}, including discriminative F1, where the baseline LLaVA-1.6-7B and PURGE-M-NPO achieve 87.0 and 93.2, respectively. Across all three benchmarks, Correct$\rightarrow$Incorrect regressions remain below 1\%, whereas Incorrect$\rightarrow$Correct transitions are substantially more frequent, indicating that PURGE corrects substantially more errors than it introduces. The remaining regressions mainly occur on visually challenging examples involving small, distant, heavily occluded, or ambiguous target objects; representative cases are provided in Appendix~\ref{app:failure_analysis}.

\section{Conclusion and limitations}
We studied spurious object-background correlations as an important source of hallucination and shortcut reliance in vision-language models. We introduced \algo, a framework that constructs structured retain/forget partitions and uses them for both controlled evaluation and partition-aware unlearning. Across multiple LVLMs, vision-language encoders, and benchmarks, \algo~reduces hallucination and improves robustness to spurious correlations while preserving competitive overall utility.
\vspace{-1mm}

\textbf{Limitations.} \algo~depends on the quality of masks, counterfactual images, cue proposals, and detector outputs. Its effectiveness may decrease when object and background cues are highly entangled, and it adds data processing and fine-tuning costs. Future work will explore more scalable partitioning and stronger counterfactual generation.

\newpage

 \section*{Acknowledgments}

We thank the anonymous reviewers for their valuable comments and suggestions. The work of Dongxiao Zhu was supported by the National Science Foundation (NSF) through Grants 2525266 and 2504264, and by the National Artificial Intelligence Research Resource (NAIRR) through project NAIRR260014. The work of Rhongho Jang was supported by the National Science Foundation (NSF) under Grant Number 2542128. The authors declare no competing interests.

\medskip
\bibliographystyle{IEEEtran}{}
\bibliography{references}


\newpage
\appendix
\textbf{Appendix}

\section{Training details}
\label{sec:TD}

\textbf{Implementation overview.} We provide detailed implementation and
training configurations for the proposed \algo~framework to ensure full
reproducibility. All experiments are conducted using parameter-efficient
fine-tuning, where only a subset of model parameters is updated while the
majority of the pretrained backbone remains frozen. Specifically, we update
the language model using LoRA adapters~\cite{hu2022lora} and, where
applicable, jointly optimize the multimodal projector while keeping the
visual encoder and remaining backbone parameters frozen. This design enables
targeted modification of both reasoning and visual grounding components while
preserving the general capabilities of the pretrained model. The language
model captures high-level semantic reasoning and is responsible for
suppressing context-driven priors, while the projector ensures correct
alignment between visual features and language representations.

\noindent\textbf{Dataset construction, splits, and size.}
We construct structured retain and forget datasets using the three partitioning strategies described in Sec.~\ref{sec:data}. For all COCO-based LVLM experiments, the candidate samples used to construct the retain and forget sets for parameter updates are drawn exclusively from the MSCOCO train2017 split. The COCO val2017 split is held out from partition-aware training and is used only for evaluation. Thus, no image from COCO val2017 enters the retain or forget sets used for model parameter updates. The partition-construction procedure on COCO train2017 produces a pool of approximately 40,000 samples spanning all 80 COCO object categories. From this pool, we select approximately 8,000 retain samples and 8,000 forget samples across the eligible object categories for each method. Thus, the approximately 40,000 samples constitute the candidate partition pool, whereas approximately 16,000 samples in total are used for parameter updates for each method. Waterbirds and other non-COCO benchmarks follow their respective dataset-specific class definitions and experimental protocols. For \algo-D, the retain and forget sets are constructed from the object-presence and object-removed counterfactual partitions described in Sec.~\ref{sec:data}. For \algo-H, retain samples correspond to consistent predictions $(1,1)$ and $(0,0)$, whereas forget samples correspond to failure cases $(1,0)$ and $(0,1)$. For \algo-M, we partition samples according to object and contextual-cue presence, using $C \cap S^c$ as the retain set and $C^c \cap S$ as the forget set. The same sampling procedure and overall training budget are used across PURGE variants to ensure a consistent comparison.

\begin{wraptable}{r}{0.30\textwidth}
    \vspace{-4mm}
    \centering
    \small
    \caption{\small Training configuration.}
    \label{tab:training_details}
    \setlength{\tabcolsep}{5pt}
    \renewcommand{\arraystretch}{1.1}
    \begin{tabular}{lc}
    \toprule
    \textbf{Parameter} & \textbf{Value} \\
    \midrule
    Number of classes & 80 \\
    Retain set size & $\sim$8,000 \\
    Forget set size & $\sim$8,000 \\
    Total samples & $\sim$16,000 \\
    Epochs & 5 \\
    Precision & bf16 \\
    Optimizer & AdamW \\
    LR (LoRA) & $2\times10^{-5}$ \\
    LR (projector) & $1\times10^{-5}$ \\
    LoRA rank & 8 \\
    $\lambda$ & 1.0 \\
    $\beta$ & 0.1 \\
    $\gamma$ & 1.0 \\
    \bottomrule
    \end{tabular}
    \vspace{-2mm}
\end{wraptable}
\textbf{Training configuration.}
We train all models for five epochs using bf16 precision with gradient checkpointing for memory efficiency. Experiments were run on a server equipped with two NVIDIA H100 PCIe GPUs, each with 80 GB of GPU memory. Optimization is performed using the AdamW optimizer~\cite{loshchilov2017decoupled}. To ensure stable optimization across components, we use separate learning rates: $2\times10^{-5}$ for LoRA parameters and $1\times10^{-5}$ for the multimodal projector. We set the LoRA rank to 8, providing a good balance between parameter efficiency and adaptation capability. For multimodal models (e.g., LLaVA and Qwen3-VL), we consider two training settings:\\
-- \one updating only LoRA adapters in the language model, and \\  -- \two jointly training LoRA together with the multimodal projector (MLP) to improve vision-language alignment. In contrast, Qwen3.5-9B does not expose a separate multimodal projector
in our training setup and is therefore adapted using LoRA only. The unlearning objective is controlled by hyperparameters $\lambda$, $\beta$, and $\gamma$. The parameter $\lambda$ determines the strength of unlearning by scaling the contribution of the forget set, and we explore different values of $\lambda$ to balance forgetting and retention. The parameter $\beta$ regularizes the KL and NPO objectives to prevent excessive deviation from the pretrained model. For \algo-D, we additionally use $\gamma$ to enforce negative supervision on object-absent samples. Unless otherwise specified, hyperparameters are kept consistent across methods and models to ensure fair comparison.

\noindent\textbf{Backbone-Specific LoRA Configuration.}
For all PURGE variants, we freeze the visual encoder and perform
parameter-efficient adaptation of the language model using LoRA.
LoRA adapters are inserted into the attention and MLP projection
modules of the language model, with rank $r=8$, scaling factor
$\alpha=16$, and dropout $0.05$. For LLaVA-1.6-7B and
Qwen3-VL-8B-Instruct, the multimodal projector/fusion component is
additionally optimized, whereas Qwen3.5-9B is adapted using LoRA only because it does not use a separate multimodal projector. All remaining backbone parameters, including the visual encoder and non-LoRA language model parameters, remain frozen. The same trainable-component configuration is used across \algo~-D,  \algo~-H, and  \algo~-M and for the GA, KL, and NPO objectives.
\vspace{-4mm}
\begin{table}[H]
\centering
\small
\caption{\small \textbf{Backbone-specific training configuration for PURGE.}
The visual encoder is frozen, while the multimodal projector/fusion
component, when present, and LLM LoRA adapters are optimized.}
\label{tab:lora_parameters}
\setlength{\tabcolsep}{3.5pt}
\renewcommand{\arraystretch}{1.08}
\begin{tabular}{lcccccc}
\toprule
\textbf{Backbone} &
\textbf{Vision} &
\textbf{Projector/Fusion} &
\textbf{LLM LoRA} &
\textbf{$r$} &
\textbf{$\alpha$} &
\textbf{Dropout} \\
\midrule
LLaVA-1.6-7B
& Frozen & Trainable & Trainable & 8 & 16 & 0.05 \\

Qwen3-VL-8B-Instruct
& Frozen & Trainable & Trainable & 8 & 16 & 0.05 \\

Qwen3.5-9B
& Frozen & N/A & Trainable & 8 & 16 & 0.05 \\
\bottomrule
\end{tabular}
\end{table}
\vspace{-4mm}





\noindent\textbf{CLIP Waterbirds Unlearning Protocol.}
For Waterbirds~\cite{sagawa2019distributionally}, we formulate the task
as binary classification between \emph{landbird} and \emph{waterbird}.
Importantly, the bird class serves as the prediction target, whereas the
background attribute is used only to characterize spurious-correlation
groups and to construct the corresponding \algo~partitions. Thus, each
sample retains its original Waterbirds class label
$y\in\{\text{landbird},\text{waterbird}\}$ throughout training.

The two bird classes crossed with the two background types
(\emph{land} and \emph{water}) define four bird-class--background
groups. These group assignments characterize whether the bird class and
background are aligned or conflicting and are used for partition
construction and worst-group evaluation; they are not treated as
classification labels. This distinction allows \algo~to suppress
background-dependent predictions while preserving bird-class
information. For CLIP-RN50, given an image $x$, the visual encoder and classification
head produce class logits $\mathbf{z}_{\theta}(x)$, with $
p_{\theta}(y\mid x)
=
\operatorname{softmax}\!\left(\mathbf{z}_{\theta}(x)\right)_y,  y\in\{\text{landbird},\text{waterbird}\}.
$ 
We instantiate the GA, KL, and NPO objectives defined in the main text
using these class probabilities and the retain/forget partitions
constructed by the corresponding \algo~strategy. In particular, the
unlearning objective acts on the bird-class probability associated with
each sample rather than predicting the four bird-class--background
groups themselves. For KL and NPO, $\theta_0$ denotes a fixed copy of
the pre-unlearning CLIP classifier and serves as the reference model.

All CLIP variants use the same CLIP-RN50 initialization. The text encoder
is frozen, while the visual encoder and classification head are optimized
using AdamW. We evaluate on the standard Waterbirds test split using
average classification accuracy and worst-group accuracy (WGA), where
WGA is the minimum classification accuracy across the four
bird-class--background groups.

\noindent\textbf{Remark.} We observe that the choice of sampling size per class provides a trade-off between computational efficiency and performance. Using 100 samples per class provides sufficient coverage while maintaining feasible training time. In preliminary experiments with smaller subsets (e.g., 30–60 samples per class), we observe consistent trends, indicating that the proposed method is robust to dataset size variations.

\section{Evaluation benchmarks}
\label{app:EB}
We evaluate our approach across multiple benchmarks that capture different aspects of hallucination and robustness. CHAIR and AMBER measure object hallucination and grounding quality in generative settings, while MMHal-Bench evaluates fine-grained multimodal reasoning. POPE focuses on object existence prediction to assess visual grounding. Causal-HalBench further examines robustness under counterfactual interventions, and MM-SpuBench evaluates sensitivity to spurious object-context correlations. Together, these benchmarks provide a comprehensive evaluation of hallucination reduction, grounding accuracy, and robustness to spurious dependencies.

\begin{table}[H]
\centering
\small
\setlength{\tabcolsep}{5pt}
\renewcommand{\arraystretch}{1.1}

\caption{ \small Evaluation benchmarks and metrics. We report benchmark-specific metrics to assess hallucination, grounding, and robustness to spurious correlations.}
\label{tab:eval_datasets_appendix}

\begin{tabular}{lll}
\toprule
\textbf{Benchmark} & \textbf{Task} & \textbf{Metrics} \\
\midrule

CHAIR~\citep{rohrbach2018object} 
& Object hallucination (generation) 
& CHAIR$_s$ $\downarrow$, CHAIR$_i$ $\downarrow$ \\

AMBER~\citep{wang2023amber} 
& Generative hallucination and grounding 
& CHAIR $\downarrow$, Cover $\uparrow$, Hal $\downarrow$, Cog $\downarrow$ \\

MMHal-Bench~\citep{sun2024aligning} 
& Fine-grained multimodal reasoning 
& Score $\uparrow$, Hallucination rate $\downarrow$ \\

POPE~\citep{li2023evaluating} 
& Object existence prediction 
& Accuracy $\uparrow$, Recall $\uparrow$, F1 $\uparrow$, Yes Ratio \\

Causal-HalBench~\citep{xu2026causal} 
& Counterfactual robustness 
& Accuracy $\uparrow$, CHR $\downarrow$, $\Delta Q$ $\downarrow$ \\

MM-SpuBench~\citep{ye2026mm} 
& Spurious correlation robustness 
& Accuracy $\uparrow$ \\

\bottomrule
\end{tabular}
\end{table}

\section{\algo-D (Inpainting-based partitioning)}
\label{app:IP}

\textbf{\algo-D-IP.} We construct the dataset using an inpainting-based strategy that explicitly disentangles object evidence from contextual cues by generating counterfactual samples (see Fig.~\ref{fig:strategies})(a). Given an input image $x$ containing a target object $o$ and background context $z$, we first obtain a binary object mask $M(x)$ that localizes the object region. Using this mask, we generate two complementary views:
$
x_r = x \odot M(x), x_f = \text{Inp}\big(x, M(x)\big),
$
where $x_r$ denotes the object-centric view obtained by removing the background and retaining only the object region, while $x_f$ denotes the context-only view obtained by removing the object and reconstructing the background via an inpainting operator $\text{Inp}(\cdot)$. Here, $\odot$ represents element-wise masking, and $\text{Inp}(\cdot)$ fills the masked object region using the surrounding visual context. Based on these views, we construct two partitions. The \textit{retain set} ${D}_r = \{(x_r, y)\}$ contains object-only samples that encourage the model to focus on object-driven features by eliminating background cues. In contrast, the \textit{forget set} ${D}_f = \{(x_f, \emptyset)\}$ consists of background-only samples where the object has been removed, providing negative supervision that discourages reliance on contextual signals.

This construction enforces an explicit separation between object and context, enabling the model to learn object-grounded representations while suppressing spurious object-context correlations. As illustrated in Fig.~\ref{fig:strategies}a, the retain and forget partitions provide complementary signals that guide the model toward robust and reliable visual grounding.

\subsection{PURGE-D Object Removal and Inpainting}
\label{app:purge_d_inpainting}

\noindent\textbf{Mask extraction and preprocessing.}
For \algo-D, we construct object-removed counterfactual images using
the instance-level annotations provided by COCO. For each target
object, the corresponding polygon/RLE segmentation annotation is
decoded into a binary object mask. We consider only non-crowd
instances and filter difficult removal cases based on object size,
proximity to the image boundary, overlap with other objects, scene
complexity, and mask solidity. Specifically, the target-object mask
must occupy $1.5\%$--$20\%$ of the image, have a maximum bounding-box
IoU of $0.08$ with other annotated objects, and have a minimum mask
solidity of $0.40$. We additionally restrict images to at most $10$
annotated objects and exclude target instances lying within $3.5\%$
of the image boundary. Before inpainting, we dilate the binary object mask using a
$13\times13$ elliptical kernel to remove residual pixels around the
object boundary. The mask is subsequently smoothed using a
$5\times5$ Gaussian filter and re-binarized. Rather than inpainting
the entire image, we extract a square local crop centered on the
target-object bounding box with $40\%$ contextual padding. The crop
size is constrained to $224$--$640$ pixels and rounded to a multiple
of eight.

\noindent\textbf{Object removal.}
We perform object removal using LaMa~\cite{Suvorov_2022_WACV} through the
\texttt{SimpleLama} implementation. The processed segmentation mask
and corresponding local image crop are supplied to the inpainter,
which reconstructs the masked target-object region from its surrounding
visual context. If the generated crop differs from the original crop
dimensions, it is resized using Lanczos interpolation before being
pasted back at the original image location. Pixels outside the local
crop remain unchanged.

\noindent\textbf{Quality control and failure handling.}
To suppress visually disruptive edits, we measure the mean absolute
RGB change in a local ring surrounding the inpainted mask boundary
and discard an edited sample when this boundary-change score exceeds
$45$. Samples with missing or empty segmentation masks, corrupted or
unreadable source images, and samples for which LaMa inpainting fails
are skipped and are not included in the constructed dataset.

\subsection{Experimental result}

\textbf{CHAIR evaluation of \algo-D-IP.}
Table~\ref{tab:chair_method1} reports CHAIR results on COCO val. All variants of {\algo-D} significantly reduce hallucination compared to the baseline across all models, demonstrating the effectiveness of inpainting-based~\cite{wang2018image} data construction. We observe a consistent improvement from GA to KL and further to NPO, indicating that stronger unlearning objectives lead to better suppression of spurious object-context dependencies. Notably, \algo-D-NPO achieves the lowest CHAIR scores, confirming improved object grounding and reduced hallucinated predictions.

\raggedbottom
\vspace{-4mm}
\begin{table}[H]
\centering
\small
\setlength{\tabcolsep}{2 pt}
\renewcommand{\arraystretch}{1}

\caption{ \small CHAIR~\cite{rohrbach2018object} evaluation on COCO val.}
\label{tab:chair_method1}

\begin{tabular}{lcccccc}
\toprule
\multirow{2}{*}{\textbf{Method}} 
& \multicolumn{2}{c}{\textbf{LLaVA-1.6-7B}} 
& \multicolumn{2}{c}{\textbf{Qwen3-VL-8B-Instruct}} 
& \multicolumn{2}{c}{\textbf{Qwen3.5-9B}} \\

& CHAIR$_s$ $\downarrow$ & CHAIR$_i$ $\downarrow$ 
& CHAIR$_s$ $\downarrow$ & CHAIR$_i$ $\downarrow$ 
& CHAIR$_s$ $\downarrow$ & CHAIR$_i$ $\downarrow$ \\
\midrule

Baseline 
& 41.2 & 23.6
& 32.4 & 19.3
& 36.2 & 21.2 \\

\midrule

\algo-D-IP-GA 
& 25.9 & 14.1
& 19.4 & 13.7
& 20.5 & 14.7 \\

\algo-D-IP-KL 
& \second{21.9} & \second{13.3}
& \second{19.3} & \second{12.1}
& \second{19.7} & \second{13.1} \\

\algo-D-IP-NPO 
& \best{18.3} & \best{12.1}
& \best{16.2} & \best{11.4}
& \best{16.6} & \best{11.9} \\

\bottomrule
\end{tabular}

\end{table}
\vspace{-4mm}

\begin{figure}[H]
    \centering
    \begin{subfigure}[t]{0.46\linewidth}
        \centering
        \includegraphics[width=\linewidth]{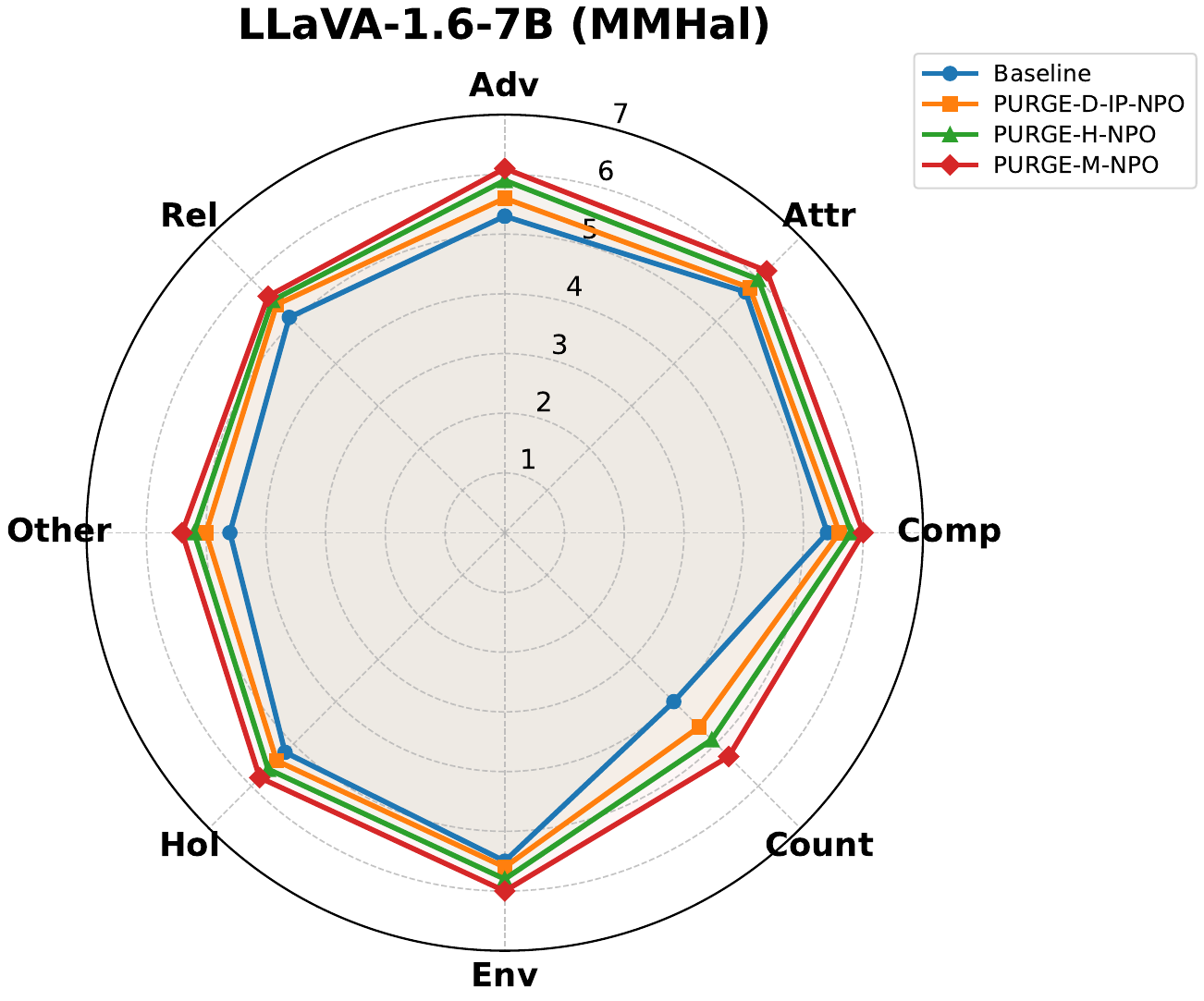}
        \caption{LLaVA-1.6}
    \end{subfigure}
    \hfill
    \begin{subfigure}[t]{0.46\linewidth}
        \centering
        \includegraphics[width=\linewidth]{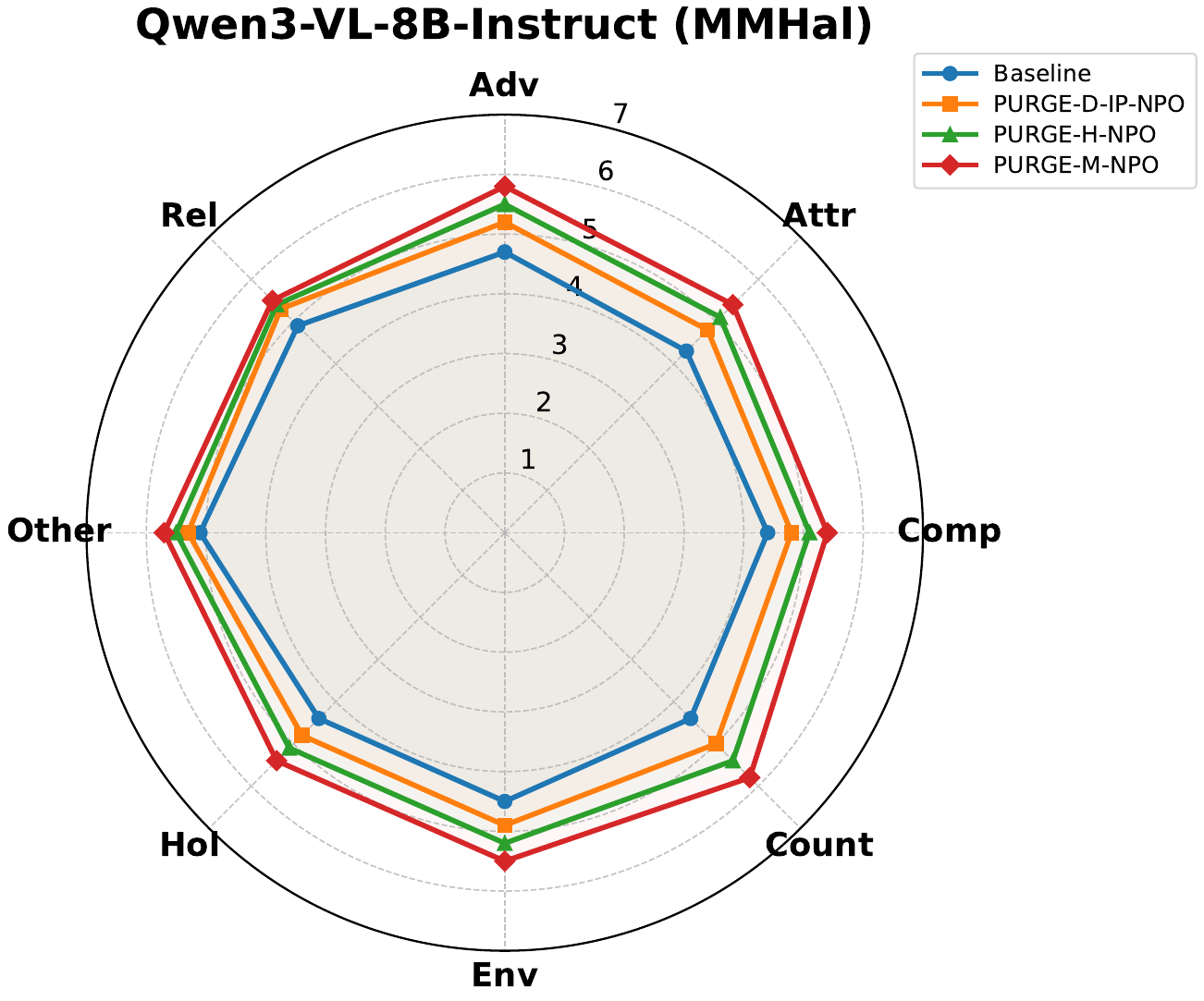}
        \caption{Qwen3-VL}
    \end{subfigure}
    \caption{MMHal~\cite{sun2024aligning} category-wise performance under \textbf{\algo-D-IP}.}
    \label{fig:mmhal_method1_inpaint_radar}
\end{figure}

\textbf{MMHal evaluation on \algo-D-IP. }Figure~\ref{fig:mmhal_method1_inpaint_radar} compares category-wise MMHal scores for LLaVA-1.6 and Qwen3-VL-8B-Instruct across the baseline, \algo-D-IP-NPO, \algo-H-NPO, and \algo-M-NPO. The IP variant is used only for \algo-D. All three PURGE methods score above the baseline in every displayed category for both backbones. Among them, \algo-M-NPO achieves the highest score in each category, followed by \algo-H-NPO and \algo-D-IP-NPO. The gains vary across categories and backbones, with clear improvements in adversarial reasoning and counting. These results show that the compared PURGE methods improve category-wise performance on MMHal, with \algo-M-NPO yielding the largest gains in these plots.

\section{Effect of forget-set ratio under NPO}
\label{app:forget_set}

To study the impact of unlearning strength, we vary the proportion of forget-set samples during training while keeping the retain set fixed. Specifically, we consider forget-set ratios of 25\%, 50\%, 75\%, and 100\%, and evaluate performance on MM-SpuBench.

\begin{figure}[H]
\centering
\includegraphics[width=0.45\textwidth]{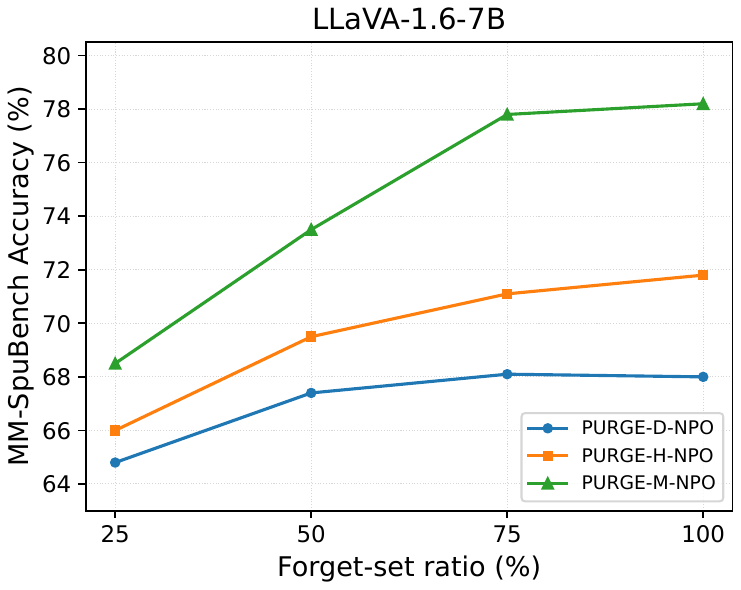}
\includegraphics[width=0.45\textwidth]{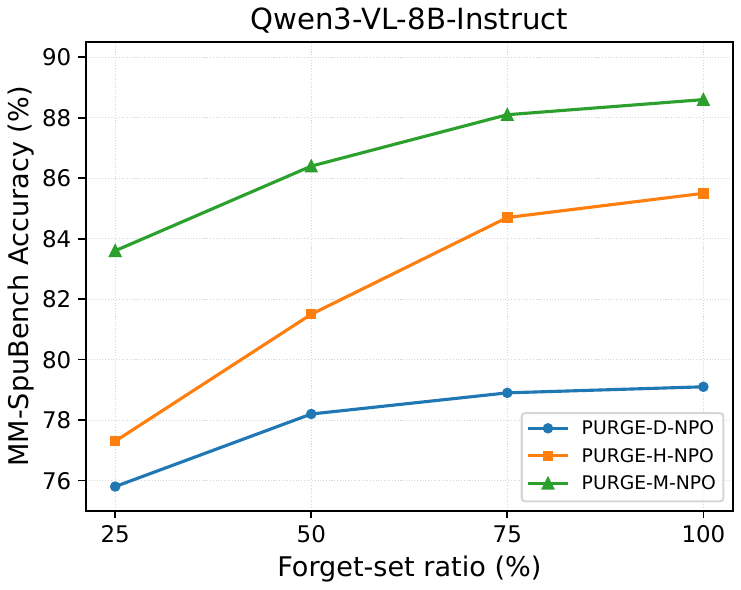}
\caption{\small {Effect of forget-set ratio under NPO.} Increasing the proportion of forget samples improves robustness to spurious correlations across both LLaVA-1.6-7B (left) and Qwen3-VL-8B-Instruct (right). \algo-M consistently achieves the best performance, followed by \algo-H and \algo-D.}
\label{fig:forget_ratio}
\end{figure}

\vspace{-4mm}
As shown in Fig.~\ref{fig:forget_ratio}, increasing the forget-set ratio consistently improves performance across all methods, indicating that stronger exposure to spurious examples helps suppress shortcut-based reasoning. However, the magnitude of improvement varies across partition strategies. \algo-D shows limited gains and saturates early, suggesting that pixel-level masking provides relatively weak supervision for disentangling object and context. In contrast, \algo-H yields steady improvements, benefiting from model-driven identification of failure cases. \algo-M consistently achieves the highest performance across all ratios, demonstrating that semantically grounded partitions targeting spurious regimes provide the most effective supervision for unlearning. We further observe that performance gains begin to plateau at higher ratios (75\%-100\%), indicating diminishing returns once sufficient spurious exposure is achieved. This trend is more pronounced for LLaVA-1.6-7B, while Qwen3-VL-8B-Instruct continues to benefit slightly, suggesting that stronger models can better leverage structured unlearning signals. Overall, these results highlight that both the \textit{quantity} and \textit{quality} of forget supervision are crucial, with partition-aware design playing a central role in effective spurious correlation removal.

\section{PURGE-M Cue Discovery and Partition Construction}
\label{app:purge_m_construction}

\noindent\textbf{Contextual-cue discovery.} We adopt the automatic spurious-cue discovery pipeline of ~\cite{hosseini2026spurlens} paper and follow its released
implementation and default cue-discovery settings. For each target
object, GPT-4 is used to propose 9 candidate contextual cues that
commonly co-occur with the object. The candidate cues are normalized and deduplicated, while unsuitable or object-overlapping cues are removed following the SpurLens filtering procedure. We retain the prompting, filtering, detector configuration, confidence-scoring, and cue-ranking procedure provided by the released SpurLens implementation, without introducing an additional manually designed cue-selection
stage. We use OWLv2~\cite{minderer2023scaling} as the open-vocabulary detector to quantify the evidence for each candidate cue in an image. For an image $v$ and candidate cue $c$, we define the cue score as
$
s(v,c)=
\max\left(
\{0\}\cup
\{q:(b,c,q)\in\mathcal{O}(v)\}
\right),
$
where $\mathcal{O}(v)$ denotes the set of OWLv2 detections, $b$ is a detected bounding box, and $q$ is its confidence score. Thus,
$s(v,c)$ corresponds to the maximum OWLv2 confidence associated with cue $c$ in image $v$. Images are ranked according to this score for each candidate cue. Following the ~\cite{hosseini2026spurlens} procedure, we compare model behavior between images exhibiting stronger and weaker evidence for each candidate cue and select the cue showing the strongest spurious
association with the target object. No additional manual cue selection is performed.

\noindent\textbf{Object--cue partition construction.}
After identifying the contextual cue, we determine target-object and cue presence for each image and construct four object--cue partitions, as visualized in Fig~\ref{fig:purge_m_visualization}. Partition membership is determined jointly by whether the target object and the selected contextual cue are present in the image.
For partition-aware unlearning, \algo-M uses images in which the target object is present but the contextual cue is absent as the retain set. These samples preserve target-object evidence while removing the identified contextual shortcut. Conversely, images in which the target object is absent but the contextual cue is present form the forget set, as they isolate contextual evidence that can spuriously support the target-object prediction. The remaining two object--cue combinations are reserved for evaluation and are not used as retain or forget samples during unlearning.

\section{Group-wise Analysis and Visualization of \algo-M}
\label{app:GA}

We evaluate \algo-M using four structured object--context partitions
that explicitly separate target-object presence from contextual-cue
presence. The first subscript denotes whether the target object is
present, while the second denotes whether the associated contextual cue
is present. Specifically, $A_{yy}$ contains samples in which both the
target object and contextual cue are present; $A_{yn}$ contains samples
in which the target object is present but the contextual cue is absent;
$A_{ny}$ contains samples in which the target object is absent but the
contextual cue is present; and $A_{nn}$ contains samples in which
neither the target object nor the contextual cue is present. For each partition, we report object-existence prediction accuracy.
Accordingly, the correct prediction is \emph{yes} for $A_{yy}$ and
$A_{yn}$, where the target object is present, and \emph{no} for
$A_{ny}$ and $A_{nn}$, where the target object is absent. Among these
partitions, $A_{ny}$ provides a direct diagnostic of contextual shortcut
reliance because the target object is absent while its associated
contextual cue remains present. An incorrect \emph{yes} prediction in
this setting indicates that the model predicts the target object from
contextual evidence alone. Table~\ref{tab:appendix_m3_all} reports group-wise object-existence accuracy on 6,332 object–context samples constructed from COCO train2017. We use these results as a training-split diagnostic to examine model behavior across the four object–context groups. The baseline models obtain relatively low accuracy on
$A_{ny}$, with scores of $32\%$, $46\%$, and $47\%$ for
LLaVA-1.6-7B, Qwen3-VL-8B-Instruct, and Qwen3.5-9B, respectively.
This behavior indicates sensitivity to contextual cues when
object-specific evidence is absent. \algo-M improves performance across
all four partitions. In particular, \algo-M-NPO improves $A_{ny}$
accuracy from $32\%$ to $59\%$ for LLaVA-1.6-7B, from $46\%$ to
$63\%$ for Qwen3-VL-8B-Instruct, and from $47\%$ to $58\%$ for
Qwen3.5-9B. Improvements are also observed on $A_{yn}$, where the
model must recognize the target object without support from the
associated contextual cue: accuracy increases from $52\%$ to $77\%$,
from $65\%$ to $80\%$, and from $62\%$ to $75\%$ for the three
backbones, respectively. These complementary improvements indicate
both reduced context-driven false-positive predictions and stronger
object-grounded recognition. Across all evaluated backbones and partitions, \algo-M-NPO also
outperforms the corresponding \algo-M-KL variant. The consistent gains
across $A_{yy}$, $A_{yn}$, $A_{ny}$, and $A_{nn}$ suggest that the
NPO objective provides a stronger forgetting signal while maintaining
performance across different object--context configurations.

\begin{figure}[H]
    \centering
    \includegraphics[width=0.90\linewidth]
    {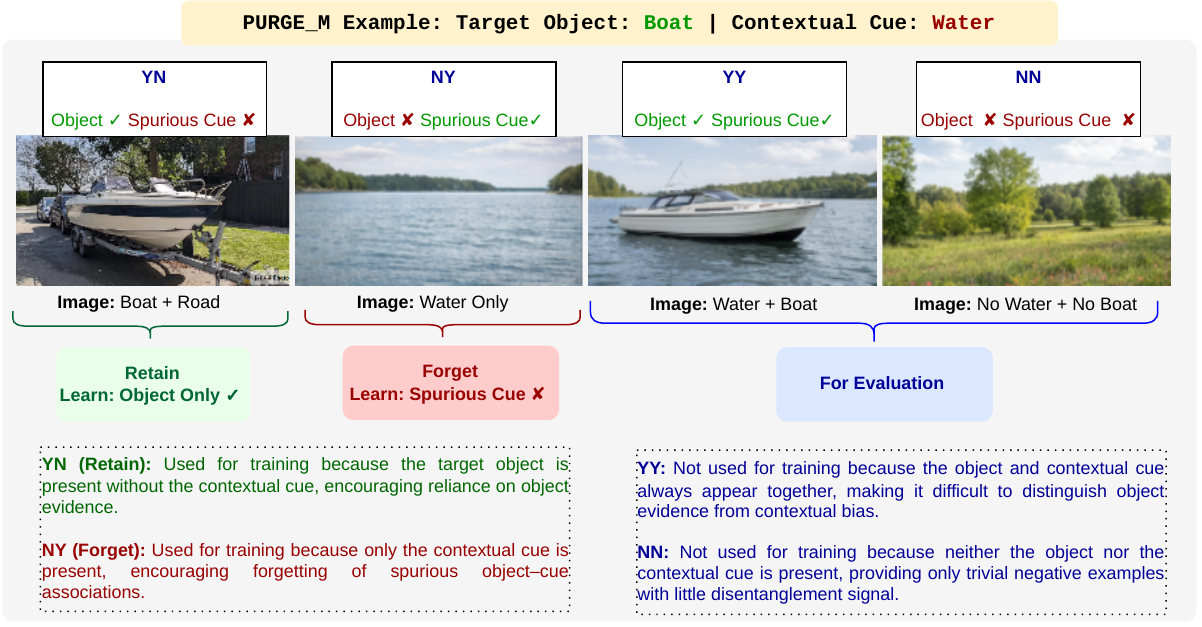}
    \caption{\textbf{Additional visualization of the PURGE-M
    partitioning strategy.}
    Example with \emph{boat} as the target object and \emph{water} as
    the contextual cue. YN (object present, cue absent) forms the
    retain set, while NY (object absent, cue present) forms the forget
    set. YY and NN are not used for unlearning and are retained for
    evaluation. The construction explicitly separates object evidence
    from contextual evidence, encouraging preservation of
    object-grounded recognition while suppressing spurious
    object--context associations.}
    \label{fig:purge_m_visualization}
\end{figure}

\vspace{-4mm}
\begin{table}[H]
\centering
\small
\setlength{\tabcolsep}{2pt}
\renewcommand{\arraystretch}{1}
\caption{\small Training-split group-wise object-existence accuracy on the four object–context partitions 
 ($A_{yy}$, $A_{yn}$, $A_{ny}$, and $A_{nn}$) constructed
from COCO train2017.}
\label{tab:appendix_m3_all}
\begin{tabular}{llcccc}
\toprule
\textbf{Model} & \textbf{Method} & $\mathbf{A_{yy} \uparrow}$ & $\mathbf{A_{yn} \uparrow}$ & $\mathbf{A_{ny} \uparrow}$ & $\mathbf{A_{nn} \uparrow}$ \\
\midrule

\multirow{3}{*}{LLaVA-1.6-7B}
& Baseline        & 60 & 52 & 32 & 45 \\
& \algo-M-KL   & 78 & 70 & 46 & 64 \\
& \algo-M-NPO  & \second{83} & \second{77} & \second{59} & \second{70} \\
\midrule

\multirow{3}{*}{Qwen3-VL-8B-Instruct}
& Baseline        & 70 & 65 & 46 & 58 \\
& \algo-M-KL   & 82 & 75 & 54 & 68 \\
& \algo-M-NPO  & \best{87} & \best{80} & \best{63} & \best{75} \\
\midrule

\multirow{3}{*}{Qwen3.5-9B}
& Baseline        & 61 & 62 & 47 & 58 \\
& \algo-M-KL   & 64 & 68 & 51 & 62 \\
& \algo-M-NPO  & 72 & 75 & 58 & 68 \\
\bottomrule
\end{tabular}
\end{table}
\vspace{-4mm}
The four constructed partitions contain 1,469 $A_{yy}$ samples,
1,578 $A_{yn}$ samples, 1,455 $A_{ny}$ samples, and
1,830 $A_{nn}$ samples, respectively. These partitions are constructed
from COCO train2017 and are used to examine model behavior under different combinations of target-object and contextual-cue presence. Figure~\ref{fig:purge_m_visualization} illustrates the partitioning
strategy using \emph{boat} as the target object and \emph{water} as
its associated contextual cue. In this example, $A_{yy}$ contains
images in which both the boat and water are present, while $A_{yn}$
contains images in which the boat is present without water.
Conversely, $A_{ny}$ contains images in which water is present but the
boat is absent, and $A_{nn}$ contains images in which neither the boat
nor water is present. For partition-aware unlearning, \algo-M uses $A_{yn}$ as the retain
set because these samples contain target-object evidence without the
associated contextual cue. In contrast, $A_{ny}$ forms the forget set
because it contains the contextual cue in the absence of the target
object and therefore isolates the shortcut signal to be suppressed.
The $A_{yy}$ and $A_{nn}$ partitions are excluded from unlearning and included in this training-split. This construction separates object-specific
evidence from contextual evidence and enables \algo-M to suppress
context-driven predictions while preserving object-grounded
recognition.

\section{Qualitative Failure Analysis}
\label{app:failure_analysis}

\begin{figure}[H]
    \centering
    \includegraphics[
       width=\textwidth,
        keepaspectratio
    ]{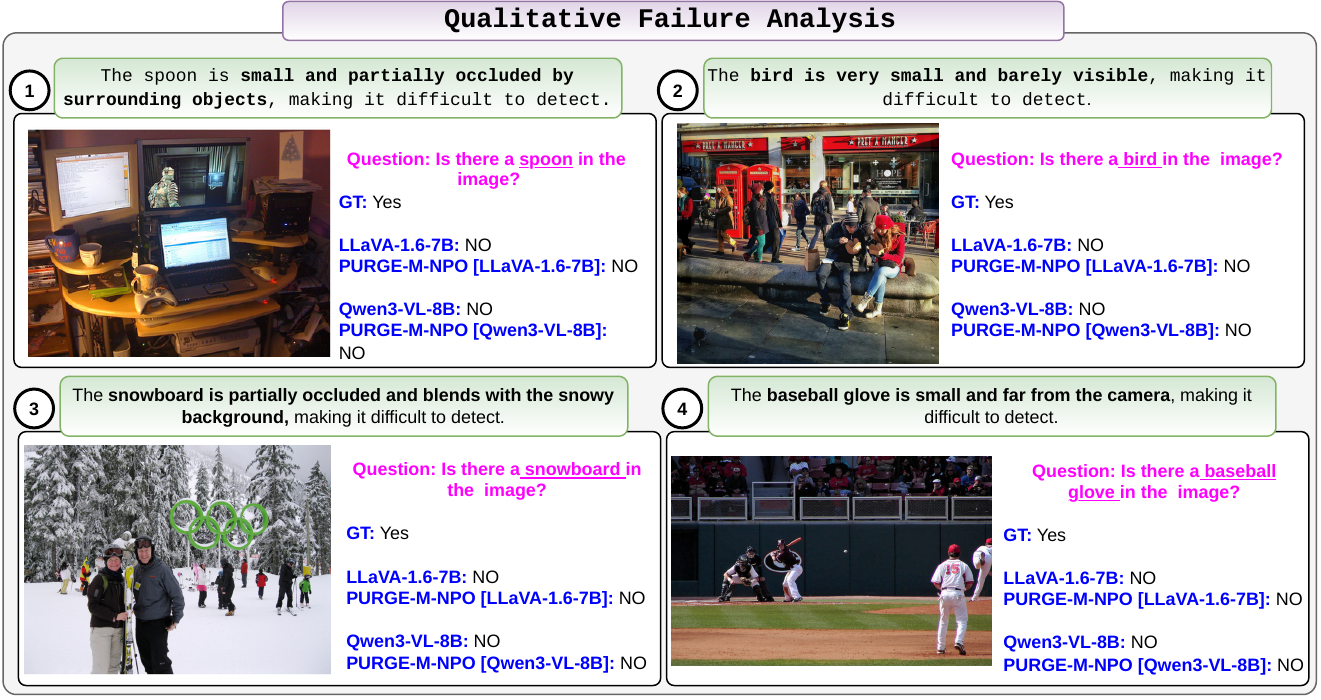}
    \caption{\textbf{Qualitative failure analysis of PURGE.}
    Representative failure cases involving small, distant, partially
    occluded, or visually ambiguous target objects. Both the baseline
    and PURGE-trained models fail in these visually challenging cases,
    illustrating that PURGE mitigates context-driven shortcut reliance
    but does not directly address failures caused by insufficient
    object-level visual evidence.}
    \label{fig:failure_cases}
\end{figure}

\vspace{-4mm}

To better characterize the remaining limitations of PURGE, we examine representative cases in which both the baseline and PURGE-trained models produce incorrect object-existence predictions. As shown in
Fig.~\ref{fig:failure_cases}, these failures primarily occur when object-specific visual evidence is weak. The target object may be very
small, distant, partially occluded, or difficult to distinguish from the surrounding scene. For example, the spoon is small and partially occluded by surrounding objects, the bird occupies only a very small image region, the snowboard is partially occluded and blends with the snowy background, and the baseball glove is small and far from the
camera. These examples clarify the scope of PURGE. The proposed framework is designed to mitigate context-driven shortcut reliance rather than all
sources of LVLM error. When object-specific visual evidence itself is insufficient or ambiguous, suppressing contextual shortcuts alone may
not recover the correct prediction.
\section{Quality Verification of Generated Partitions}
\label{app:partition_quality}

To independently assess the quality of the automatically constructed
PURGE partitions, we randomly sample 5,000 images from each partition
and verify whether they satisfy their expected object--context
configurations using Grounding DINO~\cite{liu2024grounding}, an open-vocabulary detector that
was not used during dataset construction. For PURGE-D and PURGE-H, we
verify whether the target object is present in the retain/original
images and absent from the forget/edited images. For PURGE-M, we
additionally verify the presence or absence of the associated
contextual cue. We report 95\% Wilson confidence intervals to quantify
the statistical uncertainty of the estimated verification rates.
\vspace{-4mm}
\begin{table}[H]
\centering
\small
\caption{\small \textbf{Independent verification of PURGE partition quality.}
For each PURGE partition, 5,000 examples are randomly sampled from the
larger constructed candidate partition and independently verified using
Grounding DINO.}
\label{tab:partition_verification}
\resizebox{\linewidth}{!}{
\begin{tabular}{llllc}
\toprule
Dataset & Partition & Expected Configuration &
Success Rate (\%) $\uparrow$ & 95\% Wilson CI \\
\midrule
PURGE-D & Retain & Object present
& 98.82 & [98.48, 99.08] \\
& Forget & Object absent
& 93.04 & [92.30, 93.71] \\
\midrule
PURGE-H & Original & Object present
& 97.06 & [96.55, 97.49] \\
& Edited & Object absent
& 80.08 & [78.95, 81.16]\\
\midrule
PURGE-M & YY & Object present, cue present
& 96.24 &  [95.68, 96.73] \\
& YN & Object present, cue absent
& 90.70 &  [89.86, 91.47] \\
& NY & Object absent, cue present
& 92.20 &  [91.42, 92.91] \\
& NN & Object absent, cue absent
& 92.60 &  [91.84, 93.29] \\
\bottomrule
\end{tabular}}
\end{table}
\vspace{-4 mm}
Overall, the verification results indicate that the automatically
constructed partitions largely satisfy their intended semantic configurations. The comparatively lower verification rate for the
PURGE-H edited partition reflects the additional difficulty introduced by generative inpainting and motivates treating partition quality as an explicit limitation of the framework.

\textbf{Human quality audit.} All authors participated in manually reviewing a randomly selected 10\% subset of the generated data across PURGE-D, PURGE-H, and PURGE-M. They checked whether object and contextual-cue presence matched each example’s assigned partition, whether masks removed or preserved the target object as intended, and whether inpainting removed the object while preserving plausible surrounding context. All reviewed examples met these criteria.



\section{Robustness Across Random Seeds}
\label{app:seed_robustness}
Robustness Across Random Seeds. To assess whether the observed improvements are robust to stochastic training variation, we repeat the PURGE training procedure using three independent random seeds (42, 123, and 3407). For computational efficiency, the seed-robustness experiments are conducted using a fixed reduced training subset consisting of approximately 4,000 retain and 4,000 forget samples for each run. Across seeds, we keep the sampled-data budget, training configuration, and hyperparameters fixed and vary only the random seed. We conduct this analysis for both LLaVA-1.6-7B and Qwen3-VL-8B-Instruct on MM-SpuBench, POPE, and AMBER. The resulting standard deviations remain small, indicating low sensitivity to stochastic training variation. Because this experiment uses a reduced training subset, the absolute scores in Table~\ref{tab:multiseed_all} are intended for assessing seed stability rather than direct numerical comparison with the main experimental results.

\begin{table}[H]
\centering
\small
\caption{\small \textbf{Robustness across random seeds.}
We report accuracy (\%) across three
independent runs with seeds 42, 123, and 3407. Mean and standard
deviation are computed over the three runs. Higher is better
($\uparrow$).}
\label{tab:multiseed_all}
\resizebox{\textwidth}{!}{
\begin{tabular}{lllcccc}
\toprule
Model & Benchmark & Method &
Seed 42 & Seed 123 & Seed 3407 & Mean $\pm$ Std \\
\midrule

\multirow{9}{*}{LLaVA-1.6-7B}

& \multirow{3}{*}{MM-SpuBench}
& \algo-D-NPO
& 67.54
& 68.08
& 67.79
& $67.80 \pm 0.27$ \\

&
& \algo-H-NPO
& \second{69.98}
& \second{70.41}
& \second{70.19}
& \second{$70.19 \pm 0.22$} \\

&
& \algo-M-NPO
& \best{77.96}
& \best{78.75}
& \best{78.21}
& \best{$78.31 \pm 0.40$} \\

\cmidrule(lr){2-7}

& \multirow{3}{*}{POPE}
& \algo-D-NPO
& \best{96.71}
& \second{96.55}
& \best{96.83}
& \best{$96.70 \pm 0.14$} \\

&
& \algo-H-NPO
& \second{96.54}
& \best{96.68}
& \second{96.60}
& \second{$96.61 \pm 0.07$} \\

&
& \algo-M-NPO
& 95.92
& 96.08
& 95.97
& $95.99 \pm 0.08$ \\

\cmidrule(lr){2-7}

& \multirow{3}{*}{AMBER}
& \algo-D-NPO
& \best{82.24}
& \best{82.81}
& \best{82.47}
& \best{$82.51 \pm 0.29$} \\

&
& \algo-H-NPO
& 80.74
& 81.02
& 80.91
& $80.89 \pm 0.14$ \\

&
& \algo-M-NPO
& \second{81.63}
& \second{82.02}
& \second{81.78}
& \second{$81.81 \pm 0.20$} \\

\midrule

\multirow{9}{*}{Qwen3-VL-8B}

& \multirow{3}{*}{MM-SpuBench}
& \algo-D-NPO
& 73.08
& 74.25
& 74.12
& $73.82 \pm 0.64$ \\

&
& \algo-H-NPO
& \second{78.02}
& \second{78.51}
& \best{78.29}
& \second{$78.27 \pm 0.25$} \\

&
& \algo-M-NPO
& \best{79.12}
& \best{78.92}
& \second{78.12}
& \best{$78.72 \pm 0.53$} \\

\cmidrule(lr){2-7}

& \multirow{3}{*}{POPE}
& \algo-D-NPO
& \best{98.01}
& \best{98.36}
& \best{99.12}
& \best{$98.50 \pm 0.57$} \\

&
& \algo-H-NPO
& \second{97.94}
& 97.13
& 97.28
& $97.45 \pm 0.43$ \\

&
& \algo-M-NPO
& 97.28
& \second{97.51}
& \second{97.69}
& \second{$97.49 \pm 0.21$} \\

\cmidrule(lr){2-7}

& \multirow{3}{*}{AMBER}
& \algo-D-NPO
& 87.23
& 87.60
& 87.83
& $87.55 \pm 0.30$ \\

&
& \algo-H-NPO
& \second{88.43}
& \second{88.21}
& \best{88.81}
& \second{$88.48 \pm 0.30$} \\

&
& \algo-M-NPO
& \best{88.90}
& \best{89.01}
& \second{88.72}
& \best{$88.88 \pm 0.15$} \\

\bottomrule
\end{tabular}}
\end{table}

\section{Evaluation Prompts and Answer Normalization}
\label{app:evaluation_prompts}

For completeness and reproducibility, we report the exact prompt
formats used for the major evaluation benchmarks.

\paragraph{POPE and AMBER discriminative evaluation.}
For binary object-existence questions, we use

\begin{quote}
\texttt{<image> Is there a <object> in the image?}
\end{quote}

where \texttt{<object>} denotes the queried object category.

For POPE, we use the official evaluation protocol to parse model-generated responses into binary \emph{yes}/\emph{no} predictions. The evaluator handles differences in capitalization, punctuation, and additional generated text according to the official parsing procedure. The same parser is applied to all baseline and PURGE variants across the Random, Popular, and Adversarial splits, with no method-specific manual post-processing. Accuracy, Precision, Recall, F1, and Yes ratio are then computed from the parsed predictions using the official evaluator.

\paragraph{CHAIR and AMBER generative evaluation.}
For caption generation, we use

\begin{quote}
\texttt{<image> Describe the image in detail.}
\end{quote}

\paragraph{MMHal-Bench.}
We directly use the official benchmark-provided image--question pairs
without modification:

\begin{quote}
\texttt{<image> <official benchmark question>}
\end{quote}

The official questions cover object recognition, attribute
recognition, counting, spatial reasoning, and commonsense reasoning.

\paragraph{MM-SpuBench.}
We use the official multiple-choice evaluation format:

\begin{quote}
\texttt{<image> Question: <official benchmark question>}\\
\texttt{A. <option A>}\\
\texttt{B. <option B>}\\
\texttt{C. <option C>}\\
\texttt{D. <option D>}
\end{quote}

The model is required to select the correct option from A--D.

\paragraph{Causal-HalBench.}
We directly use the official binary object-existence questions for
both original and counterfactual images:

\begin{quote}
\texttt{<image> Is there a <object> in the image?}
\end{quote}

\paragraph{Answer normalization.}
For binary evaluations, model outputs are normalized before scoring.
Responses such as \texttt{Yes}, \texttt{yes}, and \texttt{yes.} are
mapped to \texttt{Yes}, while \texttt{No}, \texttt{no}, and
\texttt{no.} are mapped to \texttt{No}. Case differences and terminal
punctuation are ignored during normalization.

\section{Object existence evaluation (POPE) across all splits}
\label{app:pope}

Table~\ref{tab:pope_all} presents full results on the POPE across Random, Popular, and Adversarial splits, which progressively increase the strength of spurious correlations. The vanilla baseline consistently underperforms across all metrics, confirming its dependence on biased object--context associations. All \algo~variants substantially improve performance, with \algo-M-NPO achieving the best overall results and \algo-H-NPO often ranking second. PURGE-M-NPO improves accuracy over the baseline on all three splits. The largest gain is on Random for LLaVA and Qwen3-VL, and on Popular for Qwen3.5; the smallest gain is on Adversarial for all three models.

\raggedbottom
\begin{table}[H]
\vspace{-4mm}
\centering
\setlength{\heavyrulewidth}{1.0pt}
\setlength{\lightrulewidth}{0.7pt}
\setlength{\cmidrulewidth}{0.7pt}

\caption{\small
POPE~\cite{li2023evaluating} results across Random, Popular, and
Adversarial splits. Best and second-best values are highlighted
in green and red, respectively.}
\label{tab:pope_all}

\resizebox{\textwidth}{!}{%
\begin{tabular}{lll|cccc|cccc|cccc}
\toprule
\multirow{2}{*}{Split}
& \multirow{2}{*}{Method}
& \multirow{2}{*}{Obj.}
& \multicolumn{4}{c|}{LLaVA-1.6-7B}
& \multicolumn{4}{c|}{Qwen3-VL-8B-Instruct}
& \multicolumn{4}{c}{Qwen3.5-9B} \\
& &
& Acc & Rec & F1 & Yes
& Acc & Rec & F1 & Yes
& Acc & Rec & F1 & Yes \\
\midrule

\multirow{10}{*}{Random}
& Baseline & --
& 67.2 & 54.5 & 62.4 & 37.3
& 79.1 & 80.7 & 79.4 & 51.6
& 79.9 & 78.8 & 79.7  & 48.9 \\

\cmidrule(lr){2-15}

& \multirow{3}{*}{\algo-D}
& GA
& 87.5 & 88.0 & 87.6 & 50.5
& 90.1 & 89.2 & 90.0 & 49.1
& 88.2 & 86.1 & 88.0 & 47.9 \\

& & KL
& 92.5 & 89.5 & 92.3 & 47.0
& 97.2 & 94.4 & 97.1 & 47.2
& 90.1 & 90.1 & 90.1 & 50.0 \\

& & NPO
& 93.7 & 89.9 & 93.5 & 46.2
& 91.1 & 92.8 & 91.2 & 51.7
& \best{92.3} & 91.6 & \best{92.2} & 49.3 \\

\cmidrule(lr){2-15}

& \multirow{3}{*}{\algo-H}
& GA
& 89.5 & 88.5 & 89.4 & 49.0
& 88.7 & 89.1 & 88.7 & 50.4
& 86.2 & 85.4 & 86.1 & 49.2 \\

& & KL
& 90.3 & 89.6 & 90.2 & 49.3
& 90.5 & 91.1 & 90.6 & 50.6
& 89.4 & 88.3 & 89.3 & 48.9 \\

& & NPO
& \second{97.3} & \second{95.1} & \second{97.2} & 47.8
& 98.1 & 96.2 & 98.1 & 48.1
& \second{92.1} & \second{91.7} & \second{92.1} & 49.6 \\

\cmidrule(lr){2-15}

& \multirow{3}{*}{\algo-M}
& GA
& 91.6 & 92.1 & 91.6 & 50.5
& 92.3 & 93.0 & 92.4 & 50.7
& 88.1 & 88.3 & 88.1 & 50.2 \\

& & KL
& 92.2 & 92.4 & 92.2 & 50.2
& \second{98.2} & \second{96.4} & \second{98.2} & 48.2
& 89.8 & 91.2 & 89.9 & 51.4 \\

& & NPO
& \best{98.1} & \best{96.2} & \best{98.1} & 48.1
& \best{99.6} & \best{99.2} & \best{99.6} & 49.6
& 91.6 & \best{92.1} & 91.6 & 50.5 \\

\midrule

\multirow{10}{*}{Popular}
& Baseline & --
& 70.0 & 54.4 & 64.5 & 34.4
& 80.1 & 81.0 & 80.3 & 50.9
& 69.9 & 78.3 & 72.2 & 58.4 \\

\cmidrule(lr){2-15}

& \multirow{3}{*}{\algo-D}
& GA
& 87.9 & 86.5 & 87.7 & 48.6
& 91.0 & 89.3 & 90.8 & 48.3
& 86.1 & 85.3 & 86.0 & 49.2 \\

& & KL
& 91.1 & 92.0 & 91.2 & 50.9
& 96.1 & 92.1 & 95.9 & 46.0
& 90.2 & 89.8 & 90.2 & 49.6 \\

& & NPO
& 92.6 & 92.0 & 92.6 & 49.4
& 96.2 & 92.5 & 96.1 & 46.3
& \second{92.0} & \second{90.4} & \second{91.9} & 48.4 \\

\cmidrule(lr){2-15}

& \multirow{3}{*}{\algo-H}
& GA
& 88.7 & 88.6 & 88.7 & 49.9
& 88.5 & 89.3 & 88.6 & 50.8
& 85.3 & 84.7 & 85.2 & 49.4 \\

& & KL
& 89.7 & 90.0 & 89.7 & 50.3
& 92.0 & 91.4 & 92.0 & 49.4
& 89.2 & 88.3 & 89.1 & 49.1 \\

& & NPO
& \second{95.2} & \second{94.0} & \second{95.1} & 48.8
& \second{96.3}  & \second{92.6} & \second{96.2} & 46.3
& \best{92.7} & 90.1 & \best{92.5} & 47.4 \\

\cmidrule(lr){2-15}

& \multirow{3}{*}{\algo-M}
& GA
& 92.7 & 92.3 & 92.7 & 49.6
& 93.0 & \second{92.7} & 93.0 & 49.7
& 85.2 & 87.1 & 85.5 & 51.9 \\

& & KL
& 92.5 & 93.1 & 92.5 & 50.6
& 92.9 & 90.6 & 92.7 & 47.7
& 88.4 & 89.1 & 88.5 & 50.7 \\

& & NPO
& \best{96.0} & \best{94.6} & \best{95.9} & 48.6
& \best{97.5} & \best{95.0} & \best{97.4} & 47.5
& 90.1 & \best{91.3} & 90.2 & 51.2 \\

\midrule

\multirow{10}{*}{Adversarial}
& Baseline & --
& 77.6 & 55.2 & 71.1 & 27.6
& 87.9 & 79.0 & 86.7 & 41.1
& 89.7 & 79.3 & 88.5 & 39.6 \\

\cmidrule(lr){2-15}

& \multirow{3}{*}{\algo-D}
& GA
& 90.4 & 84.2 & 89.8 & 43.8
& 95.1 & 90.3 & 94.9 & 45.2
& 88.9 & 82.1 & 88.1 & 43.2 \\

& & KL
& 89.6 & 87.3 & 89.4 & 47.7
& 92.3 & 89.0 & 92.0 & 46.7
& 93.1 & 88.7 & 92.8 & 45.6 \\

& & NPO
& 92.9 & 90.8 & 92.7 & 47.9
& 94.6 & 93.1 & 94.5 & 48.5
& 93.9 & \second{90.1} & 93.7 & 46.2 \\

\cmidrule(lr){2-15}

& \multirow{3}{*}{\algo-H}
& GA
& 91.9 & 88.2 & 91.6 & 46.3
& 93.3 & 89.1 & 93.0 & 45.8
& 91.0 & 85.2 & 90.4 & 44.2 \\

& & KL
& 91.8 & 89.7 & 91.6 & 47.9
& 94.1 & 91.1 & 93.9 & 47.0
& 93.0 & 88.2 & 92.6 & 45.2 \\

& & NPO
& \best{96.2} & \best{95.1} & \best{96.2} & 48.9
& 94.3 & 88.7 & 94.0 & 44.4
& \best{95.3} & \best{91.6} & \best{95.1} & 46.3 \\

\cmidrule(lr){2-15}

& \multirow{3}{*}{\algo-M}
& GA
& \second{94.6} & 89.3 & \second{94.3} & 44.7
& 94.1 & 92.1 & 94.0 & 48.0
& 91.6 & 84.1 & 90.9 & 42.5 \\

& & KL
& 92.8 & 90.6 & 92.6 & 47.8
& 96.9 & \second{93.8} & \second{96.8} & 46.9
& 93.1 & 86.2 & 92.6 & 43.1 \\

& & NPO
& 93.0 & \second{91.7} & 92.9 & 48.7
& \second{97.1} & \best{94.2} & \second{97.0} & 47.1
& \second{94.7} & 89.4 & \second{94.4} & 44.7 \\

\bottomrule
\end{tabular}%
}
\end{table}

\section{Object hallucination reduction on CHAIR}
\label{app:chair}
Table~\ref{tab:chair_main} evaluates object hallucination using the CHAIR metric on the COCO validation set. 
Baseline models exhibit high hallucination rates, as reflected by large CHAIR$_s$ and CHAIR$_i$ values, indicating frequent generation of non-existent objects. Our proposed {\algo}~framework consistently reduces hallucinations across all models and settings. 
In particular, the {\algo-M-NPO} variant achieves the best performance, significantly lowering CHAIR scores compared to both baseline and other variants. 
For example, on LLaVA-1.6-7B, CHAIR$_s$ is reduced from 41.2 to 11.3, demonstrating a substantial improvement in object grounding.  Moreover, the consistent gains across different architectures highlight
the effectiveness of partition-aware unlearning in mitigating spurious
correlations, which are an important contributor to hallucination.

\begin{table}[H]
\centering
\small
\setlength{\tabcolsep}{4pt}
\renewcommand{\arraystretch}{1}
\caption{\small {CHAIR~\cite{rohrbach2018object} evaluation on COCO validation set.} 
Lower is better.}
\label{tab:chair_main}

\begin{tabular}{lcccccc}
\toprule
\multirow{2}{*}{\textbf{Method}} 
& \multicolumn{2}{c}{\textbf{LLaVA-1.6-7B}} 
& \multicolumn{2}{c}{\textbf{Qwen3-VL-8B-Instruct}} 
& \multicolumn{2}{c}{\textbf{Qwen3.5-9B}} \\

& CHAIR$_s \downarrow$ & CHAIR$_i \downarrow$ 
& CHAIR$_s \downarrow$ & CHAIR$_i \downarrow$ 
& CHAIR$_s \downarrow$ & CHAIR$_i \downarrow$ \\
\midrule

\textbf{Baseline} 
& 41.2 & 23.6
& 32.4 & 19.3
& 36.2 & 21.2 \\

\midrule

\multicolumn{7}{c}{\textbf{\algo-D}} \\
\algo-D-GA 
& 26.9 & 15.3 
& 21.2 & 13.5
& 30.6 & 19.6 \\

\algo-D-KL 
& 24.5 & 14.1 
& 20.3 & 13.2 
& 27.1 & 18.5 \\

\algo-D-NPO 
& 20.3 & 12.7
& 17.4 & 11.9
& 24.8 & 18.9 \\

\midrule

\multicolumn{7}{c}{\textbf{\algo-H}} \\
\algo-H-GA 
& 22.5 & 17.6
& 19.3 & 12.4 
& 25.8 & 19.1 \\

\algo-H-KL 
& 19.9 & 15.7 
& 17.4 & 11.1 
& 24.6 & 18.3 \\

\algo-H-NPO 
& 18.3 & \second{9.3} 
& \second{11.6} & \second{7.4} 
& 23.3 & 17.0 \\

\midrule

\multicolumn{7}{c}{\textbf{\algo-M}} \\
\algo-M-GA 
& 20.4 & 15.1 
& 18.2 & 11.7 
& 22.6 & 16.3 \\

\algo-M-KL 
& \second{17.5} & 11.2 
& 16.1 & 9.2 
& \second{21.6} & \second{16.0} \\

\algo-M-NPO 
& \best{11.3} & \best{6.6}
& \best{9.6} & \best{4.3}
& \best{20.7} & \best{14.2} \\

\bottomrule
\end{tabular}
\end{table}

\section{Fine-grained evaluation on MM-SpuBench}
\label{app:MM}

\begin{wrapfigure}{r}{0.36\textwidth}
    \vspace{-4mm}
    \centering
    \includegraphics[width=\linewidth]{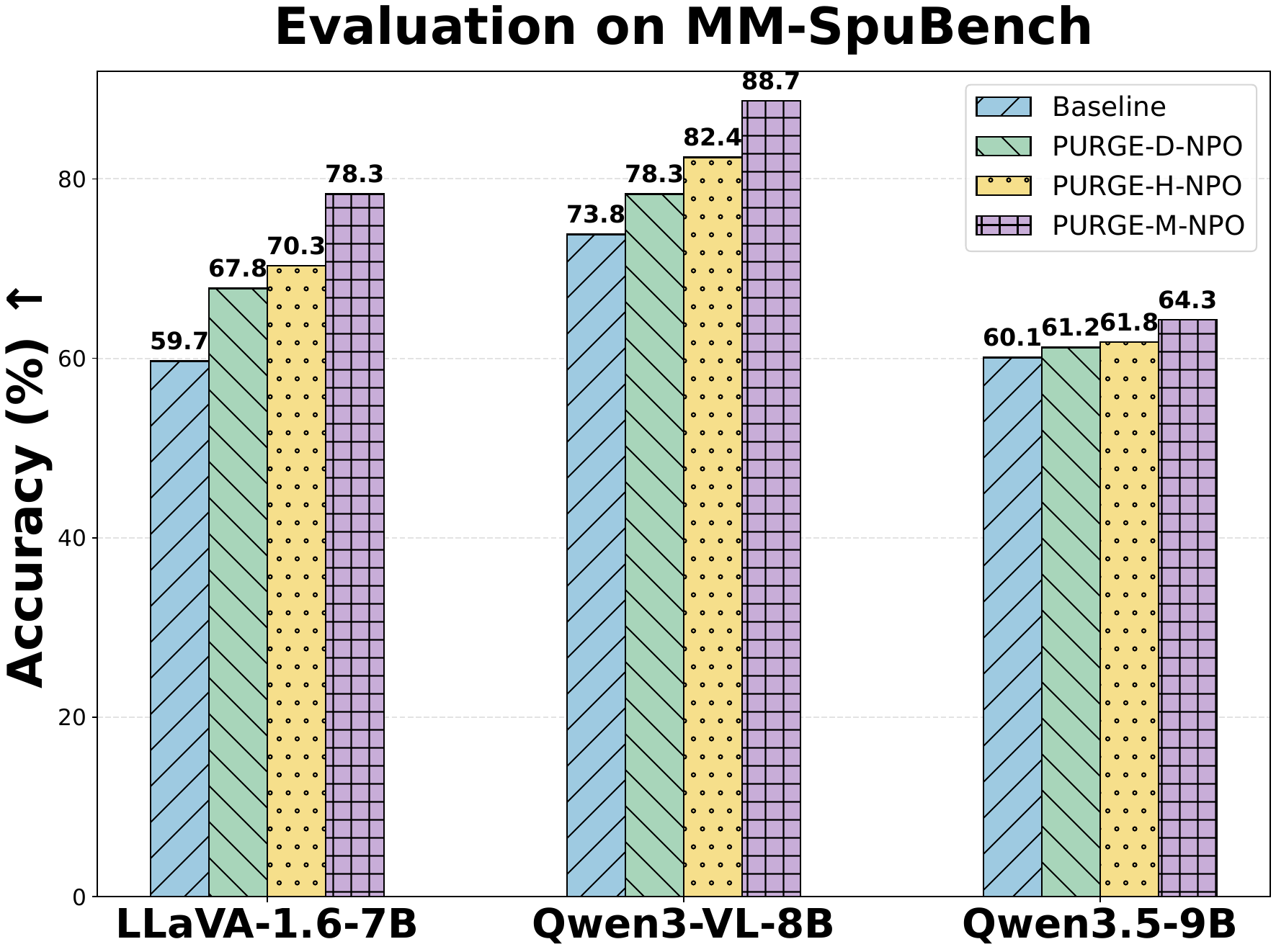}
    \caption{\small Evaluation on MM-SpuBench. \algo~consistently improves 
    accuracy across models, with the largest gains under \algo-M-NPO.}
    \label{fig:mmspu}
    \vspace{-4mm}
\end{wrapfigure}
Fig.~\ref{fig:mmspu} evaluates performance on MM-SpuBench~\cite{ye2026mm}. All \algo~variants consistently improve accuracy across models, with \algo-M achieving the largest gains, improving LLaVA-1.6-7B from 59.7 to 78.3 and Qwen3-VL-8B-Instruct from 73.8 to 88.7, demonstrating effective removal of spurious correlations. Table~\ref{tab:mmspubench_category} presents category-wise performance across multiple spurious correlation dimensions. Baseline models exhibit strong bias, particularly in categories such as BG and RS, indicating over-reliance on contextual cues. Our {\algo}~framework consistently improves performance across all categories, with the largest gains observed in bias-sensitive metrics. Among variants, {\algo-M-NPO} achieves the best results across all models, significantly improving both individual categories and overall accuracy. These improvements highlight the effectiveness of partition-aware unlearning in mitigating spurious correlations and enhancing generalization.



\begin{table}[H]
\centering
\caption{\textbf{Category-wise accuracy (\%) on MM-SpuBench.}
All category metrics are higher-is-better ($\uparrow$).
Macro Avg denotes the unweighted mean accuracy across the nine spurious-correlation categories. Overall MM-SpuBench accuracy is reported separately in Table~\ref{tab:ab_MM}.}
\label{tab:mmspubench_category}

\resizebox{\textwidth}{!}{
\begin{tabular}{llcccccccccc}
\toprule
\textbf{Model}
& \textbf{Method}
& \textbf{BG$\uparrow$}
& \textbf{TN$\uparrow$}
& \textbf{CO$\uparrow$}
& \textbf{RS$\uparrow$}
& \textbf{Col$\uparrow$}
& \textbf{Ori$\uparrow$}
& \textbf{LS$\uparrow$}
& \textbf{PA$\uparrow$}
& \textbf{Sha$\uparrow$}
& \textbf{ Macro Avg$\uparrow$}
\\
\midrule

\multirow{4}{*}{LLaVA-1.6-7B}
& Baseline
& 67.5
& 52.7
& 61.7
& 73.8
& 43.0
& 70.1
& 50.9
& 44.5
& 59.5
& 58.2
\\

& PURGE-D-NPO
& 70.5
& 56.8
& 65.4
& 75.3
& 46.4
& 72.3
& 54.2
& 48.3
& 63.1
& 61.4
\\

& PURGE-H-NPO
& 74.1
& 60.2
& 68.2
& 78.3
& 50.6
& 75.5
& 58.7
& 52.7
& 67.3
& 65.1
\\

& PURGE-M-NPO
& 79.4
& 66.2
& 73.6
& 82.1
& 56.3
& 80.2
& 64.3
& 58.3
& 73.1
& 70.4
\\

\midrule

\multirow{4}{*}{Qwen3-VL-8B-Instruct}
& Baseline
& 82.9
& 51.3
& 70.6
& 84.7
& 59.4
& 74.8
& 72.4
& 66.0
& 66.2
& 69.8
\\

& PURGE-D-NPO
& 85.2
& 59.0
& 73.0
& 86.0
& 63.3
& 77.5
& 75.2
& 70.2
& 70.8
& 73.4
\\

& PURGE-H-NPO
& \second{87.0}
& \second{64.0}
& \second{77.0}
& \second{88.0}
& \second{67.4}
& \second{81.0}
& \second{79.8}
& \second{75.1}
& \second{75.3}
& \second{77.2}
\\

& PURGE-M-NPO
& \best{90.0}
& \best{71.5}
& \best{82.6}
& \best{91.0}
& \best{73.8}
& \best{85.0}
& \best{84.3}
& \best{81.6}
& \best{81.2}
& \best{82.3}
\\

\midrule

\multirow{4}{*}{Qwen3.5-9B}
& Baseline
& 60.5
& 50.3
& 61.2
& 72.4
& 42.0
& 68.0
& 52.1
& 46.5
& 60.2
& 57.0
\\

& PURGE-D-NPO
& 61.6
& 51.4
& 62.3
& 73.3
& 43.1
& 69.3
& 53.3
& 47.8
& 61.1
& 58.1
\\

& PURGE-H-NPO
& 62.4
& 52.8
& 63.8
& 74.0
& 44.8
& 70.0
& 54.1
& 48.5
& 62.4
& 59.2
\\

& PURGE-M-NPO
& 64.3
& 55.6
& 65.1
& 75.0
& 47.5
& 72.4
& 56.0
& 50.7
& 64.8
& 61.3
\\

\bottomrule
\end{tabular}
}
\end{table}

\section{Robustness and hallucination evaluation}
\label{app:amber}

\textbf{AMBER benchmark: hallucination and grounding.} Table~\ref{tab:amber_main} evaluates hallucination and grounding performance across generative and discriminative settings. 
Baseline models exhibit high hallucination (CH, Hal) and lower coverage, indicating poor object grounding and reliance on spurious correlations.  Table~\ref{tab:amber_main} shows mixed results for PURGE-D-NPO: it improves several metrics, but LLaVA’s F1E decreases from 95.1 to 93.5 and F1R decreases from 69.6 to 57.2. PURGE-M-NPO improves every reported metric across the evaluated models. In particular, {\algo-M-NPO} achieves the best results, significantly reducing hallucination while improving coverage and discriminative scores. 
For example, on Qwen3-VL-8B-Instruct, CH reduces from 5.9 to 2.8 and Hal from 26.5 to 8.3, demonstrating strong improvements in factual consistency and object grounding.

\textbf{Ablation study on AMBER.}
Table~\ref{tab:amber_capacity} analyzes the effect of different components of our method. 
We observe that partition-aware unlearning alone (LLM) already reduces hallucination compared to the baseline, confirming the effectiveness of removing spurious dependencies. Further improvements are achieved by incorporating the projector (LLM + Proj), which enhances visual alignment and leads to lower hallucination and higher coverage. These results indicate that both unlearning and improved vision-language alignment are complementary and essential for achieving robust and reliable predictions.

\begin{table}[H]
\centering
\small
\setlength{\tabcolsep}{4pt}
\renewcommand{\arraystretch}{1.05}
\caption{\small {AMBER evaluation.}}
\label{tab:amber_main}

\resizebox{\textwidth}{!}{%
\begin{tabular}{l l cccc cccc}
\toprule
\multirow{2}{*}{Model} & \multirow{2}{*}{Method}
& \multicolumn{4}{c}{Generative}
& \multicolumn{4}{c}{Discriminative} \\
\cmidrule(lr){3-6} \cmidrule(lr){7-10}
& & CH$\downarrow$ & Cov$\uparrow$ & Hal$\downarrow$ & Cog$\downarrow$
& F1$\uparrow$ &  F1E$\uparrow$ &  F1A$\uparrow$ & F1R$\uparrow$ \\
\midrule

\multirow{4}{*}{LLaVA-1.6-7B}
& Baseline       & 8.3 & 61.0 & 48.6 & 4.2 & 87.0 & 95.1 & 81.5 & 69.6 \\
& \algo-D-NPO  & 6.3 & 65.6 & 23.7 & 3.8 & 90.2 & 93.5 & 87.4 & 57.2 \\
& \algo-H-NPO & 4.3 & 68.3 & 16.3 & 3.0 & 90.8 & 95.5 & 88.7 & 72.1 \\
& \algo-M-NPO  & \second{3.1} & 74.1 & \second{11.3} & 2.7 & \second{93.2} & \best{96.8} & \second{90.0} & 78.4 \\
\midrule

\multirow{4}{*}{\shortstack{Qwen3-VL-8B-\\Instruct}}
& Baseline       & 5.9 & 62.3 & 26.5 & 3.1 & 85.4 & 94.2 & 82.5 & 70.8 \\
& \algo-D-NPO  & 5.0 & 72.5 & 21.6 & 2.8 & 87.2 & 94.5 & 85.2 & 77.6 \\
& \algo-H-NPO & 4.1 & \second{79.6} & 16.3 & \second{2.1} & 92.5 & 95.3 & 87.8 & \second{80.0} \\
& \algo-M-NPO  & \best{2.8} & \best{85.2} & \best{8.3} & \best{1.1} & \best{94.6} & \second{96.7} & \best{90.1} & \best{85.2} \\
\midrule

\multirow{4}{*}{Qwen3.5-9B}
& Baseline       & 7.8 & 55.8 & 47.3 & 4.5 & 81.2 & 90.6 & 85.4 & 69.2 \\
& \algo-D-NPO  & 6.6 & 58.3 & 29.4 & 3.9 & 84.1 & 91.3 & 87.6 & 72.7 \\
& \algo-H-NPO & 5.3 & 65.4 & 22.4 & 3.7 & 85.1 & 91.8 & 88.3 & 74.2 \\
& \algo-M-NPO  & 4.4 & 68.2 & 21.6 & 3.2 & 87.2 & 92.4 & 88.6 & 78.5 \\
\bottomrule
\end{tabular}%
}
\end{table}

\begin{table}[H]
\centering
\small
\setlength{\tabcolsep}{3.2pt}
\renewcommand{\arraystretch}{1.08}
\caption{\small {Full ablation on AMBER benchmark.}}
\label{tab:amber_capacity}

\begin{tabular}{l|cccc|cccc|cccc}
\toprule
\textbf{Method}
& \multicolumn{4}{c|}{\textbf{LLaVA-1.6-7B}}
& \multicolumn{4}{c|}{\textbf{Qwen3-VL-8B-Instruct}}
& \multicolumn{4}{c}{\textbf{Qwen3.5-9B}} \\

& CH$\downarrow$ & Cov$\uparrow$ & Hal$\downarrow$ & Cog$\downarrow$
& CH$\downarrow$ & Cov$\uparrow$ & Hal$\downarrow$ & Cog$\downarrow$
& CH$\downarrow$ & Cov$\uparrow$ & Hal$\downarrow$ & Cog$\downarrow$ \\
\midrule

\textbf{Baseline}
& 8.3 & 61.0 & 48.6 & 4.2
& 5.9 & 62.3 & 26.5 & 3.1
& 7.8 & 55.8 & 47.3 & 4.5 \\
\midrule

\multicolumn{13}{c}{\textbf{\algo-D-NPO}} \\
LLM
& 7.3 & 62.1 & 39.3 & 4.0
& 5.1 & 63.7 & 25.7 & 3.0
& 6.6 & 58.3 & 29.4 & 3.9 \\

LLM + Proj
& 6.3 & 65.6 & 23.7 & 3.8
& 5.0 & 72.5 & 21.6 & 2.8
& - & - & - & - \\
\midrule

\multicolumn{13}{c}{\textbf{\algo-H-NPO}} \\
LLM
& 5.8 & 63.0 & 30.0 & 3.8
& 4.6 & 73.1 & 20.2 & 2.5
& 5.3 & 65.4 & 22.4 & 3.7 \\

LLM + Proj
& 4.3 & 68.3 & 16.3 & 3.0
& 4.1 & \second{79.6} & 16.3 & \second{2.1}
& - & - & - & - \\
\midrule

\multicolumn{13}{c}{\textbf{\algo-M-NPO}} \\
LLM
& 4.0 & 66.8 & 26.3 & 3.0
& 4.2 & 75.1 & 19.4 & 2.3
& 4.4 & 68.2 & 21.6 & 3.2 \\

LLM + Proj
& \second{3.1} & 74.1 & \second{11.3} & 2.7
& \best{2.8}& \best{85.2} & \best{8.3} & \best{1.1}
& - & - & - & - \\

\bottomrule
\end{tabular}
\end{table}

\section{Comparison of PURGE Partitioning Strategies}
\label{app:strategy_comparison}

PURGE-D, PURGE-H, and PURGE-M derive their partitions from different
sources of evidence: pixel-level object--background separation,
model behavior on original and counterfactual images, and semantic
object--context cues, respectively. Their performance varies across
evaluation settings. As shown in Table~\ref{tab:strategy_performance},
PURGE-D achieves the highest accuracy on POPE-Random, PURGE-H achieves
the highest F1 on POPE-Popular and accuracy on POPE-Adversarial, and
PURGE-M achieves the highest worst-group accuracy on Waterbirds and
accuracy on MM-SpuBench.
\vspace{-4mm}

\begin{table}[H]
\centering
\small
\setlength{\tabcolsep}{5pt}
\caption{Performance of PURGE-D, PURGE-H, and PURGE-M across
representative evaluation settings. LVLM results use NPO; the MM-SpuBench result uses the LLM+Proj. setting.}
\label{tab:strategy_performance}
\begin{tabular}{lllccc}
\toprule
Benchmark & Model & Metric
& PURGE-D & PURGE-H & PURGE-M \\
\midrule
Waterbirds
& CLIP-RN50
& WGA $\uparrow$
& 75.1 & 79.3 & \textbf{79.6} \\

POPE-Random
& Qwen3.5-9B
& Accuracy $\uparrow$
& \textbf{92.3} & 92.1 & 91.6 \\

POPE-Popular
& Qwen3.5-9B
& F1 $\uparrow$
& 91.9 & \textbf{92.5} & 90.2 \\

POPE-Adversarial
& LLaVA-1.6-7B
& Accuracy $\uparrow$
& 92.9 & \textbf{96.2} & 93.0 \\

MM-SpuBench
& Qwen3-VL-8B-Instruct
& Accuracy $\uparrow$
& 78.3 & 82.4 & \textbf{88.7} \\
\bottomrule
\end{tabular}
\end{table}
\vspace{-4mm}
The strategies also differ in construction time, as shown in
Table~\ref{tab:strategy_cost}. PURGE-D with masking has the lowest construction cost, PURGE-H has intermediate cost, and PURGE-M requires the most time because it includes semantic cue discovery and detection. Together, these results show that each strategy offers a different balance between evaluation performance and construction cost.
\vspace{-4mm}
\begin{table}[H]
\centering
\small
\setlength{\tabcolsep}{5pt}
\caption{Approximate partition-construction time using two H100 GPUs.}
\label{tab:strategy_cost}
\begin{tabular}{lcc}
\toprule
Strategy & MSCOCO & Waterbirds \\
\midrule
PURGE-D (masking)
& 4 h 20 min & 25 min \\

PURGE-D (inpainting)
& 5 h 39 min & 1 h 10 min \\

PURGE-H
& 9 h 56 min & 1 h 45 min \\

PURGE-M
& 16 h 55 min & 2 h 55 min \\
\bottomrule
\end{tabular}
\end{table}
\vspace{-4mm}
\noindent
PURGE-D is fastest; PURGE-H performs best on selected POPE splits. PURGE-M leads on Waterbirds and MM-SpuBench, at higher construction cost.

\newpage
\section*{NeurIPS Paper Checklist}


\begin{enumerate}

\item {\bf Claims}
    \item[] Question: Do the main claims made in the abstract and introduction accurately reflect the paper's contributions and scope?
    \item[] Answer:\answerYes{} 
    \item[] Justification: The abstract and introduction clearly state the main contribution: a partition-aware unlearning framework (PURGE) for mitigating spurious object–background correlations in LVLMs. The claims regarding hallucination reduction and improved robustness are supported by empirical evaluations across multiple models and benchmarks.

\item {\bf Limitations}
    \item[] Question: Does the paper discuss the limitations of the work performed by the authors?
    \item[] Answer: \answerYes{} 
    \item[] Justification: The limitations of the proposed approach are discussed in the paper.

\item {\bf Theory assumptions and proofs}
    \item[] Question: For each theoretical result, does the paper provide the full set of assumptions and a complete (and correct) proof?
    \item[] Answer: \answerNA{}
    \item[] Justification: The paper does not present formal theoretical results or proofs. The contribution is primarily methodological and empirical.

    \item {\bf Experimental result reproducibility}
    \item[] Question: Does the paper fully disclose all the information needed to reproduce the main experimental results of the paper to the extent that it affects the main claims and/or conclusions of the paper (regardless of whether the code and data are provided or not)?
    \item[] Answer: \answerYes{} 
    \item[] Justification: We provide detailed descriptions of dataset construction, training procedures, model configurations, and evaluation benchmarks. Key components such as partition definitions, loss functions, and hyperparameters are explicitly specified, enabling reproducibility.

\item {\bf Open access to data and code}
    \item[] Question: Does the paper provide open access to the data and code, with sufficient instructions to faithfully reproduce the main experimental results, as described in supplemental material?
    \item[] Answer: \answerYes{} 
    \item[] Justification: The dataset and code are publicly available. Access information is provided in the paper.

\item {\bf Experimental setting/details}
    \item[] Question: Does the paper specify all the training and test details (e.g., data splits, hyperparameters, how they were chosen, type of optimizer) necessary to understand the results?
    \item[] Answer: \answerYes{} 
    \item[] Justification: We describe model architectures, datasets, training configurations, hyperparameters, and evaluation protocols. Details include optimizer settings, learning rates, LoRA configuration, and dataset sizes.

\item {\bf Experiment statistical significance}
    \item[] Question: Does the paper report error bars suitably and correctly defined or other appropriate information about the statistical significance of the experiments?
    \item[] Answer: \answerNo{} 
    \item[] {Justification: We repeated training with three random seeds for two models using a smaller training dataset (4,000 retain and 4,000 forget examples). We report the mean and standard deviation for tests of misleading visual context (MM-SpuBench), object-presence answers (POPE), and hallucination and visual grounding (AMBER). However, the main results are single runs. The confidence intervals in the appendix assess partition quality, not variation in benchmark scores.}

\item {\bf Experiments compute resources}
    \item[] Question: For each experiment, does the paper provide sufficient information on the computer resources (type of compute workers, memory, time of execution) needed to reproduce the experiments?
    \item[] Answer: \answerYes{} 
    \item[] Justification: The paper reports the training setup, parameter-efficient tuning, and compute hardware (two NVIDIA H100 PCIe GPUs with 80 GB of GPU memory each). Experiment runtimes are not reported.
    
\item {\bf Code of ethics}
    \item[] Question: Does the research conducted in the paper conform, in every respect, with the NeurIPS Code of Ethics \url{https://neurips.cc/public/EthicsGuidelines}?
    \item[] Answer: \answerYes{} 
    \item[] Justification: The research adheres to NeurIPS ethical guidelines. The work focuses on improving the reliability and robustness of LVLMs by reducing hallucination and spurious correlations, which contribute positively to safe deployment.

\item {\bf Broader impacts}
    \item[] Question: Does the paper discuss both potential positive societal impacts and negative societal impacts of the work performed?
    \item[] Answer: \answerYes{} 
    \item[] Justification: We discuss the positive societal impacts in the introduction.

\item {\bf Safeguards}
    \item[] Question: Does the paper describe safeguards that have been put in place for responsible release of data or models that have a high risk for misuse (e.g., pre-trained language models, image generators, or scraped datasets)?
    \item[] Answer: \answerNA{} 
    \item[] Justification: The paper does not release sensitive models or datasets that pose significant misuse risks beyond standard pretrained models.

\item {\bf Licenses for existing assets}
    \item[] Question: Are the creators or original owners of assets (e.g., code, data, models), used in the paper, properly credited and are the license and terms of use explicitly mentioned and properly respected?
    \item[] Answer: \answerYes{} 
    \item[] Justification: We use publicly available datasets (e.g., MSCOCO, Waterbirds) and pretrained models (e.g., LLaVA, Qwen), all of which are properly cited and widely used in prior work.

\item {\bf New assets}
    \item[] Question: Are new assets introduced in the paper well documented and is the documentation provided alongside the assets?
    \item[] Answer: \answerYes{} 
    \item[] Justification: The derived datasets are documented in the Hugging Face dataset card and Croissant file. The dataset-generation code is publicly available, and access information is provided in the paper.
    

\item {\bf Crowdsourcing and research with human subjects}
    \item[] Question: For crowdsourcing experiments and research with human subjects, does the paper include the full text of instructions given to participants and screenshots, if applicable, as well as details about compensation (if any)? 
    \item[] Answer: \answerNA{} 
    \item[] Justification: The paper does not involve human subjects or crowdsourced data collection.

\item {\bf Institutional review board (IRB) approvals or equivalent for research with human subjects}
    \item[] Question: Does the paper describe potential risks incurred by study participants, whether such risks were disclosed to the subjects, and whether Institutional Review Board (IRB) approvals (or an equivalent approval/review based on the requirements of your country or institution) were obtained?
    \item[] Answer: \answerNA{} 
    \item[] Justification: No human subject data or experiments are involved.

\item {\bf Declaration of LLM usage}
    \item[] Question: Does the paper describe the usage of LLMs if it is an important, original, or non-standard component of the core methods in this research? Note that if the LLM is used only for writing, editing, or formatting purposes and does \emph{not} impact the core methodology, scientific rigor, or originality of the research, declaration is not required.
    \item[] Answer: \answerYes{} 
    \item[] Justification: LLMs are used as components of the evaluated models. In addition, GPT-4 is used to propose candidate contextual cues for PURGE-M during dataset construction.

\end{enumerate}

\end{document}